\documentclass[sn-basic]{sn-jnl}
\usepackage{graphicx}%
\usepackage{multirow}%
\usepackage{amsmath,amssymb,amsfonts}%
\usepackage{amsthm}%
\usepackage{mathrsfs}%
\usepackage[title]{appendix}%
\usepackage{xcolor}%
\usepackage{textcomp}%
\usepackage{manyfoot}%
\usepackage{booktabs}%
\usepackage{algorithm}%
\usepackage{algorithmicx}%
\usepackage{algpseudocode}%
\usepackage{listings}%
\usepackage{tabularx}
\usepackage{soul} 
\usepackage{adjustbox}
\usepackage{makecell}
\usepackage{pdfpages}

\usepackage{hyperref}
\usepackage{enotez}
\usepackage{xr}

\let\footnote=\endnote

\lstdefinestyle{promptstyle}{
  basicstyle=\ttfamily\small,
  breaklines=true,
  breakatwhitespace=false,
  columns=fullflexible,
  keepspaces=true,
  showstringspaces=false,
  frame=single,
  rulecolor=\color{black!20},
  backgroundcolor=\color{black!3},
  numbers=left,
  numberstyle=\tiny\color{black!50},
  xleftmargin=1.5em,
  framexleftmargin=1em,
  tabsize=2
}

\usepackage{fvextra}
\usepackage{tabularx}
\usepackage[table]{xcolor}

\DefineVerbatimEnvironment{promptblock}{Verbatim}{fontsize=\small,breaklines=true,commandchars=\\\{\},frame=single}

\definecolor{sigcol}{HTML}{2E86AB}

\usepackage{xurl}

\usepackage[commandnameprefix=ifneeded, todonotes={textsize=small}]{changes}

\definechangesauthor[color=magenta]{ga}
\definechangesauthor[color=teal]{bea}
\definechangesauthor[color=violet]{mars}
\definechangesauthor[color=brown]{ar}
\definechangesauthor[color=blue]{sc}
\definechangesauthor[color=red]{mn}
\definechangesauthor[color=blue]{og}

\theoremstyle{thmstyleone}%
\theoremstyle{thmstyletwo}%

\theoremstyle{thmstylethree}%

\begin{document}

\title[Article Title]{Generative AI Practices, Literacy, and Divides:\\ 
An Empirical Analysis in the Italian Context}

\author*[1]{\fnm{Beatrice} \sur{Savoldi}}\email{bsavoldi@fbk.eu}
\author[2]{\fnm{Giuseppe} \sur{Attanasio}}
\author[3]{\fnm{Olga} \sur{Gorodetskaya}}
\author[4]{\fnm{Marta} \sur{Marchiori Manerba}}
\author[5]{\fnm{Elisa} \sur{Bassignana}}
\author[6]{\fnm{Silvia} \sur{Casola}}
\author[1]{\fnm{Matteo} \sur{Negri}}
\author[7]{\fnm{Tommaso} \sur{Caselli}}
\author[1]{\fnm{Luisa} \sur{Bentivogli}}
\author[1]{\fnm{Alan} \sur{Ramponi}}
\author[8]{\fnm{Arianna} \sur{Muti}}
\author[8]{\fnm{Nicoletta} \sur{Balbo}}
\author[8]{\fnm{Debora} \sur{Nozza}}

\abstract{
The rise of generative AI (GenAI) chatbots accessible via conversational interfaces is transforming digital interactions and holds economic promise. However, these tools might deepen existing inequalities---not only through uneven, socially stratified adoption, but through differentials in their purposeful, critical use. Drawing on original survey data from 1,906 Italian-speaking adults, we provide a comprehensive analysis of GenAI adoption, literacy, and usage patterns. Our findings show that GenAI is supporting diversified personal and professional activities and replacing traditional information-seeking tools. Yet less-educated and older individuals, and those with lower technology familiarity, are less likely to adopt it; 40\% cite competence barriers as a key obstacle. Among users, AI training emerges as the primary predictor of purposeful, capital-enhancing engagement---content creation, learning, and creativity enhancement---while more passive, recreational uses (e.g., companionship, information seeking) remain insensitive to competence levels. We thus highlight digital literacy as a lever for how people leverage GenAI, not just whether they use it. Finally, gender operates as a persistent cross-cutting divide, shaping both adoption and usage frequency. These findings challenge the assumption that high accessibility translates into broadly shared gains. Rather, they offer a granular, multi-level account of emerging disparities in the GenAI era---with implications for how this technology may ultimately drive outcomes and benefit divides.}


\maketitle

\section{Introduction}\label{sec:intro}

The rapid advancement of AI language technologies (LTs)---computational models that process and generate human language---has entered a transformative phase, with powerful ``GenAI chatbots'' such as ChatGPT \citep{openai2023chatgpt} reshaping how we work, access information, and interact with technology.
Unlike prior LTs designed for a dedicated activity or modality, these new tools are general-purpose, capable of performing numerous tasks---from writing and translation to coding and tutoring---across languages without domain-specific training. This makes them uniquely accessible to diverse users and adaptable across contexts. 
As a result, GenAI chatbots have achieved unprecedented adoption: By 2025, the ChatGPT platform had become one of the five most visited websites in the world \citep{explodingtopics2025websites}, 
leaving researchers, policymakers, and developers to grapple with fundamental questions about adoption, usage, and global impact. 

As a new commodity, GenAI chatbots offer promises for personalized learning \citep{abbes2024generative}, creativity \citep{holzner2025generative}, and workforce productivity \citep{noy2023experimental,brynjolfsson2025generative}.
The impact of access---or lack thereof---can be substantial: a temporary ChatGPT ban in Italy led to a significant drop in coding productivity, with output levels recovering only after users found strategies to circumvent the restrictions \citep{kreitmeir2023unintended}.

Despite widespread accessibility, emergent research suggests that GenAI might exacerbate existing inequalities and \textit{digital divides} \citep{SUAREZ2025102997, capraro2024impact}, that is, differences between those who do and do not benefit from digital tooling, thereby reinforcing social and economic marginalization in knowledge economies \citep{van2005deepening, hargittai2018digital}.
Recent evidence points to higher early uptake concentrated among men, more educated individuals \citep{hartley2024labor}, and higher-income regions \citep{daepp2025emerginggenerativeartificialintelligence}. Beyond adoption, divides can emerge in usage and skill levels, which mediate the critical and effective use of technologies \citep{unesco2024ailiteracy, bentley2024digital, wang2025artificial}. 
 Indeed, GenAI carries well-documented limitations---including the spread of misinformation \citep{chen2024combating}, sycophantic agreement \citep{batista2026rationalanalysiseffectssycophantic}, and biased, stereotyped portrayals of social groups \citep{mitchell2025shades}---that threaten reliability and may be consequential for users who lack the competencies to detect them. 
 Such risks disproportionately harm vulnerable groups. Prior research has linked algorithmic bias to economic disadvantages for women \citep{savoldi-etal-2024-harm}, while lower levels of education and digital proficiency have been associated with greater susceptibility to online scams \citep{MOUNCEY2025100125}

Recent research has examined whether and how GenAI may relate to existing inequalities. 
However, current studies lack a comprehensive, multi-layered empirical account of divides in adoption and their relationship to usage and skills. Existing inquiries focus on adoption or usage frequency only, often within constrained domains---e.g., coding \citep{prather2024wideninggapbenefitsharms}, academic writing \citep{tang2025gender}---or specific subgroups (e.g., students \citep{sublime2024chatgpt}, workers \citep{hartley2024labor}), leaving both the broader population and the full range of GenAI use cases underexplored.
Conversely, analyses of large-scale user conversations that document emerging applications of GenAI chatbots rely on proprietary, English-speaking data from single platforms (i.e., Microsoft Bing Copilot, ChatGPT, or Claude \citep{handa2025economic,chatterji2025people, tomlinson2025workingaimeasuringapplicability}) and lack reported sociodemographic information. Moreover, such studies inherently focus only on current users, overlooking those who do not engage with these tools---a potentially distinct group whose constraints and needs remain insufficiently understood \citep{zhou2025attentionnonadopters}.

In this paper, we address this gap by focusing on the Italian context.
%
%
Italy ranks 23rd out of 27 EU Member States in the Digital Economy and Society Index (DESI) for basic digital skills,\footnote{\url{https://digital-decade-desi.digital-strategy.ec.europa.eu/datasets/desi/charts}} and lags behind for levels of digital innovation.\footnote{\url{https://en.ilsole24ore.com/art/european-sme-digitisation-grows-but-italy-lags-behind-AI4CJSX}}
With respect to AI specifically, Ipsos' 2025 AI Monitor \citep{ipsos2025aimonitor} reveals that among Italians who have heard of AI, only half report truly understanding it---placing Italy 29th out of 30 countries ranked. Also, these figures conceal a stark gender gap: national statistics \citep{istat}  document a persistent divide in digital skills favouring men, which compounds with age,
and has been linked to gender gaps in employment \citep{gui2011digital, martinez2017digital, perifanou2020gender, lopez2021digital}.
Hence, Italy offers a compelling lens for examining 
whether GenAI can broaden access to digital opportunities or whether it is more likely to reproduce existing inequalities, while also providing informative insights for more ``average'' countries.

To this end, we present  a \textbf{comprehensive} \textbf{inquiry of GenAI chatbot adoption, usage patterns, and literacy in Italy}. Drawing on newly collected survey data ($N = 1,906$) that include both adopters and non-adopters alongside rich demographic information, we combine descriptive and quantitative statistical analyses with qualitative examination of open-ended responses to explore whether GenAI chatbots 
reflect digital divides in adoption (first-order divide), and in the ability and modes of use (second-order divide). We further explore the role of prior AI-based LTs familiarity, training and literacy, and compare adoption trends with those of related and pre-existing technologies to provide insights into GenAI chatbots' role within the current technological ecosystem. In conclusion, we elaborate on how GenAI could shape the third-order digital divide (i.e., social differentials in the benefits and outcomes of using this new technology), and offer suggestions for future research. 

Our study makes four key contributions. 
First, we provide a multi-level empirical account of GenAI divides within a single analytical framework. 
Second, we extend existing research beyond English-speaking contexts by focusing on Italy, which represents a critical use case. 
Third, we survey and include non-adopters, a group underexplored in existing research. 
Finally, we make all data publicly available, providing a rich snapshot of the Italian context in 2025 to enable longitudinal monitoring. 

\section{Background}\label{sec:related}

GenAI chatbots represent the latest milestone in LTs for information and communication. Powered by general-purpose Large Language Models (LLMs),\footnote{We use the term GenAI chatbots to distinguish between the core technology (i.e., LLMs) and the tools available in conversational interfaces accessible to the broader population.}  GenAI chatbots mark a leap from earlier task-specific solutions. Capable of handling diverse tasks across languages and modalities, and delivered through intuitive, user-friendly conversational interfaces, they hypothetically place powerful capabilities within reach of a broad population.
Yet emerging research documents heterogeneity in their uptake, potentially reflecting the uneven unlocking of their functionalities. To understand these dynamics, we frame our empirical study within a multilevel concept of the digital divide, with a  focus on the adoption and subsequent use of GenAI chatbots.

\subsection{First-order divide: Adoption}

The classic notion of the digital divide focuses on physical access to connectivity infrastructure and devices \citep{oecd2001digital}. The digital divide literature, however, has long argued that inequalities operate at multiple levels---from foundational questions of who accesses technology to deeper disparities in abilities and in how it is used, and in the benefits it yields \citep{hargittai2002second, dimaggio2004digital, van2006digital, SCHEERDER20171607}.  Specifically to the GenAI chatbots landscape---where systems are rapidly deployed and Western internet penetration is high---physical access is insufficient as an analytical category \citep{carter2020exploring, wang2025artificial}.\footnote{\citet{eurostat2025} data from 2025 indicate that 94.12\% of Italian households have internet access.}

Thus, the first-order divide can be understood more expansively as a willingness divide---differences in attitudes and readiness to adopt GenAI that go beyond traditional connectivity and device ownership, and that are salient in high-access European societies.
This reconfiguration is supported by evidence that technology engagement is shaped not only by material conditions but also by motivational dimensions---including beliefs about technology outcomes, perceived utility, and self-efficacy \citep{afroogh2024trust}. Along these lines, established frameworks such as the Technology Acceptance Model (TAM) \citep{davis1989technology} identify perceived usability and utility as drivers of behavioral intention to adopt. Recent work repurposing the model has found that individuals who perceive GenAI as both useful and manageable are more likely to integrate it into their daily practices \citep{khoso2026trust}.

These perceptions are likely socially stratified, and might reflect both deliberate or involuntary disinterest in technological waves \citep{wyatt2002they}. 
Beliefs about productivity gains, problem-solving capabilities or economic returns in the workplace can drive utility \citep{capraro2024impact}, which resonate less with older individuals. 
On the other hand, social norms, institutional contexts, and the gendered meanings attached to technology can influence women's behavioural intentions \citep{wajcman2004techno}:  stereotypes framing technology as a male domain may reduce identification and perceived relevance, whereas unequal distribution of domestic work can limit time available to explore new tools \citep{10.1108/DTS-04-2025-0083}. 
Specifically in the Italian context, national statistics indicate that around 50\% of men report feeling at ease with digital technologies, compared to only 15\% of women \citep{dondena2020}.
Ultimately, we hypothesize that prior technological exposure acts as a source of cumulative advantage---individuals with experience in other LTs are more likely to perceive GenAI as accessible and useful, consistent with evidence that regional-level GenAI diffusion is conditioned by a territory's pre-existing familiarity with related technologies \citep{VICENTE2025103030}.


With this reconfiguration in mind, we explore GenAI chatbot adoption as a first site of digital divides, examining the role of sociodemographic factors and prior experience with related LTs as potential sources of cumulative advantage, alongside the motivational reasons underlying (un)willingness to adopt.

\subsection{Second-Order Divide: Usage and Literacy}

Second-order divides capture inequality that persists through differences in skills and usage patterns. Indeed, \citet{hargittai2002beyond} demonstrates that Internet users vary substantially in their ability to locate, evaluate, and strategically leverage information online. Crucially, such disparities influence not only whether generic engagement with a technology is sustained, but how it is used and what benefits it derives. Accordingly, we suggest exploring GenAI uptake by examining both the nature of its use and the literacy levels that may shape engagement.

\paragraph{Usage}

GenAI chatbots support a wide and rapidly evolving range of activities, spanning professional tasks \citep{tomlinson2025workingaimeasuringapplicability}, writing and software development \citep{handa2025economic}, and everyday information-seeking, with personal and recreational usage growing faster than professional ones \citep{chatterji2025people}. However, such fine-grained evidence derives from large-scale proprietary user logs, whereas research on differential usage \citep{otis2024global} focuses on specific application areas and investigates second-order divides based on interaction volume, i.e., frequency of usage.   
Yet the nature of use may be as consequential as its volume. \citet{bassignana-etal-2025-ai} find that users of lower socioeconomic status report higher engagement with GenAI chatbots for entertainment, compared to higher-status users%
---echoing \citet{hargittai2008digital}'s distinction between general online engagement and capital-enhancing activities. Gender has also emerged as a key factor shaping usage volume across occupations \citep{otis2024global}, echoing a longstanding gendered divide in technology use \citep{toupin2024shaping, barry2025gendered}. Among French and Italian students, males consistently report higher usage rates across disciplines \citep{sublime2024chatgpt, mogelvang2024gender}. Along this line, in Norway, female students engage with GenAI chatbots less frequently, and express a stronger need to learn appropriate use \citep{mogelvang2024gender}. 
However, how literacy and digital competence more broadly shape modes of GenAI chatbots use---rather than merely volumes---remains underexplored.


\paragraph{Literacy}
Though no conclusive definition of GenAI literacy exists \citep{rapanta2025critical}, it can be understood as 
the ability to understand, critically engage with, and effectively use AI technologies
across personal and professional contexts \citep{long2020}. 
Although GenAI chatbots are widely accessible to the non-technical public, 
functional and critical literacy on their potential and limitations can mediate effective interactions and risk avoidance \citep{bouyzourn2025shapes}. 
Public concern is already significant: a global study finds that 54\% of Italians have experienced errors from GenAI and 84\% support regulatory measures against AI-generated misinformation \citep{gillespie2025trust}, with similar concerns expressed in the United States \citep{pewresearch2025google}. 
Beyond risk avoidance, prior work on digital divides has shown that literacy  influences both the efficacy and purposefulness of engagement \citep{hargittai2008digital, dimaggio2004digital}: lower-literacy users may rely on GenAI for passive, recreational uses, whereas 
higher-competence users might engage in capital-enhancing activities with greater workplace value. 

Literacy is learnable but 
it is not merely a technical competence---it is a socially structured resource shaped by education, cultural capital, and access to learning environments \citep{adam1995feminist, adam2006artificial,martini2025digital}. 
Under this assumption, individuals with higher status and education are often at an advantage, as they are more likely to have received formal training or developed competence through exposure to related technologies. 
%
Gendered perceptions of competence are also consequential: women were found to report lower self-assessed technological ability than men, a gap that does not always reflect actual performance differences \citep{hargittaiwomen} but nonetheless might shape the extent and nature of their digital engagement.

Taken together, the second-order divide in the GenAI context is not simply about volume of use, but about the nature and value of engagement---and the literacy that underpins it. With this in mind, we aim to explore a more granular account of the second-order divide. Specifically, we examine usage and its frequency across different intent categories and whether these patterns differ across literacy levels, AI education, and their sociodemographic correlates. 

\section{Design and Methods}\label{sec:method}

We provide here a brief outline of our study design, followed by detailed information about our sample, measures, and analytical methods.

\subsection{Study Design}

Our survey design is organized intro three main sections as follows. 

First, we measure adoption rates of GenAI chatbots alongside five longstanding task-specific LT applications: machine translation (e.g., Google Translate), voice assistants (e.g., Siri), assisted writing (e.g., Grammarly), speech transcription (e.g., Zoom captions), and text-to-speech generation (e.g., Adobe Read Out Loud). These applications were selected for their widespread commercial use and allow us to assess \textit{i)}~participants' familiarity and experience with related LTs, \textit{ii)} compare adoption rates to frame GenAI chatbots' relevance in the current technological ecosystem. For each technology that participants reported using, we asked whether usage had changed since the introduction of GenAI chatbots and about the reasons for this shift. Additionally, we explicitly ask about potential behavioral shifts in the use of Web search queries. 

Second, to assess literacy and competence in AI language technologies, we collected information on prior training, self-reported knowledge and competence, awareness of these technologies' benefits and risks, and desiderata for future education.

Third, for participants who adopted GenAI chatbots (henceforth GenAI \textit{users}), we asked questions mapping how, when, and for what purposes they use these tools. We provided a list of 24 \textit{activities} identified as significant in prior research \citep{ouyang-etal-2023-shifted} (e.g., email writing, coding, travel planning), asked participants to indicate whether they perform these activities for work or personal use, and inquired about 
the language (e.g., Italian or English) used in such interactions. 
While activities provide fine-grained behavioral detail, for broader insights into the mode of usage we also collected data on interaction \textit{intents}, i.e., higher-level purposes that cut across specific activities\footnote{Intents are more encompassing than individual activities, and allow us to capture patterns beyond the specific activities currently attested in the literature. This is relevant to include unanticipated uses that the rapidly evolving GenAI landscape may not yet have fully documented.} (see also Section \ref{subsec:measures}). We rely on established taxonomies of usage intent \citep{bodonhelyi2024userintentrecognitionsatisfaction, 10.1145/3732294}.


Finally, we collected sociodemographic information, including age, gender, geography, socioeconomic status, and educational level.\footnote{We also collect information on the professional area of the participants, which is released with the data. However, focus on professions are deemed beyond the scope of the current study and hence not discussed.}

To ensure accessibility for non-expert participants,  we follow \citet{bassignana-etal-2025-ai} and provide a friendly definition of LT (i.e., computational models that process and generate human language), distinguishing between general-purpose GenAI chatbots and task-specific LTs, alongside examples and descriptions for each of our five specific LT applications. 
The survey---distributed in Italian and translated here for dissemination purposes---was extensively piloted and refined to enhance clarity, construct validity, and reliability. We provide the full list of definitions (Table~\ref{tab:definitions} in Section~\ref{appA:definitions}) and chatbot activities (Section~\ref{appA:activities}) in the Supplementary materials. Table \ref{tab:intent} in the main body reports the intent categories.

\begin{table}[t]
\centering
\label{tab:user_intents}
\begin{tabular}{p{2.5cm}p{4.5cm}p{4.5cm}}
\toprule
\textbf{User Intent} & \textbf{Description} & \textbf{Examples} \\
\midrule
Information Retrieval 
& Requesting factual information or instructions.
& \textit{What is the capital of Italy?} \\
\addlinespace

Problem Solving 
& Support in managing tasks or decisions.
& \textit{My PC is slow, how can I fix it?} \\
\addlinespace

Learning 
& Support in acquiring new skills or understanding concepts.
& \textit{Explain the difference between Hegel's and Kant's theories.} \\
\addlinespace

Content Creation 
& Generating, editing, or revising materials.
& \textit{Make the following text shorter and more persuasive.} \\
\addlinespace

Personal Leisure 
& Questions asked for fun or companionship.
& \textit{What is your name?} \\
\addlinespace

Creativity 
& Seeking ideas, inspiration, or suggestions.
& \textit{Give me 5 Christmas gift ideas.} \\
\bottomrule
\end{tabular}
\caption{\textbf{Intent categories and definitions for GenAI chatbot usage.} The taxonomy of intents is drawn by the five categories outlined by~\citet{10.1145/3732294}, with the addition of the ``Creativity'' intent validated by~\citet{bodonhelyi2024userintentrecognitionsatisfaction}. For each of the six intents, we present the descriptions and one of the three examples provided in the survey.}
\label{tab:intent}
\end{table}

\begin{table}[t]
\centering
\begin{tabular}{@{}lccc@{}}
\toprule
\textbf{Variable} & \textbf{GenAI Users} & \textbf{Non-Users} & \textbf{Total} \\
& $n = 1,533$ & $n = 373$ & $N = 1,906$ \\
\midrule
\textbf{Gender (\%)} & & & \\
\quad Men & 810 (52.8) & 126 (33.8) & 936 (49.1) \\
\quad Women & 700 (45.7) & 237 (63.5) & 937 (49.2) \\
\quad Non-binary/Other & 23 (1.5) & 10 (2.7) & 33 (1.7) \\
\addlinespace
\textbf{Age (\%)} & & & \\
\quad 18--34 & 491 (32.0) & 44 (11.8) & 535 (28.1) \\
\quad 35--54 & 498 (32.5) & 85 (22.8) & 583 (30.6) \\
\quad 55--64 & 304 (19.8) & 92 (24.7) & 396 (20.8) \\
\quad 65+ & 240 (15.7) & 152 (40.8) & 392 (20.6) \\
\addlinespace
\textbf{Education (\%)} & & & \\
\quad Graduates & 1,120 (73.1) & 202 (54.2) & 1,322 (69.4) \\
\quad Non-graduates & 413 (26.9) & 171 (45.8) & 584 (30.6) \\
\addlinespace
\textbf{Geography (\%)} & & & \\
\quad North & 917 (59.8) & 237 (63.5) & 1,154 (60.6) \\
\quad Centre & 253 (16.5) & 58 (15.6) & 311 (16.3) \\
\quad South and Islands & 271 (17.7) & 66 (17.7) & 337 (17.7) \\
\quad Abroad & 92 (6.0) & 12 (3.2) & 104 (5.5) \\
\addlinespace
\textbf{Socioeconomic Status (\%)} & & & \\
\quad Lower & 281 (18.3) & 79 (21.2) & 361 (18.9) \\
\quad Mid & 802 (52.3) & 220 (59.0) & 1,020 (53.5) \\
\quad Higher & 450 (29.4) & 74 (19.8) & 525 (27.5) \\
\bottomrule
\end{tabular}
\caption{\textbf{Description of the sample}. We provide distributions for the full sample (\emph{Total}; $N = 1,906$) as well as disaggregated across participants who reported using GenAI chatbots in the last twelve months (\emph{GenAI Users; $n = 1,533$}) and those that did not (\emph{Non-Users; $n = 373$}).}
\label{tab:sample_description}
\end{table}

\subsection{Participants and Distribution}
The questionnaire was distributed between May 23rd and August 27th, 2025, using the Qualtrics platform \citep{qualtrics2025}. We use a non-probability sampling strategy \citep{vehovar2016non}, where respondents voluntarily self-select into the survey. 
To ensure broad coverage and approximate the Italian population more closely, the survey was disseminated through multiple online channels, including social media, mailing lists, and messaging apps such as WhatsApp and Telegram. Broader outreach was achieved via national and local media outlets, including \emph{Corriere della Sera},\footnote{\url{https://www.corriere.it/}} \emph{TP24},\footnote{\url{https://www.tp24.it/} } and \emph{Il Fatto Quotidiano}.\footnote{\url{https://www.ilfattoquotidiano.it/}} Participants were encouraged to share the survey within their networks, fostering a snowball sampling effect. They completed the survey anonymously on a device of their choice.

Eligibility required participants to be at least 18 years old and have native or  C1 level proficiency in Italian (standard CEFR language proficiency scale). Responses failing this language filter, representing 4.55\% of submissions ($n=100$), were excluded. Additional filters were applied to enhance data quality. We removed responses completed in less than 2.78 minutes ($n=3$)---below the lower interquartile range of the median survey duration of 8.7 minutes---as well as responses flagged by Qualtrics as potential duplicates ($n=66$). Participants with missing responses on the  GenAI chatbot adoption question ($n=10$) were also excluded.
After applying these criteria, the final analytical sample consisted of 1,906 respondents. Key sample characteristics are summarized in Table~\ref{tab:sample_description}. We report age, gender, geography (including Italian people living abroad), education, and socioeconomic status. This was operationalized by asking respondents ``In which socioeconomic bracket do you place yourself?" The responses were then recoded into three categories: Lower (``Low'' and ``Middle-low''), Mid (``Middle''), and Higher (combining ``Mid-high" and "High"). We proportionally redistributed ``I prefer not to answer" responses (2.6\%) across the three categories. 

Our sample demonstrates a reasonable alignment with the key demographic characteristics of the Italian population (comparison with national statistics in Supplementary Section~\ref{appAsample}, based on Italian National Statistics Institute data---i.e., ISTAT 2025\footnote{\url{https://demo.istat.it/app/?i=POS&l=it}}).
Gender distribution is balanced and closely mirrors national statistics. Age shows a distribution typical of online surveys, with a slight underrepresentation (65+ years old; 20.06\% in our sample vs. 28.7\% in the general population). 
Educational attainment shows the largest deviation from a representative sample (30.6\% of non-graduates, vs. 85\% in the general population), also due to a higher representation of younger individuals in our sample. On the one hand, this outcome aligns with established work proving that education modulates web survey populations \citep{bethlehem2010selection,hargittai2018biases}. On the other hand, extensive literature on digital divides establishes that higher educational attainment is among the primary predictors of digital skills and use \citep{deursen2010measuring,blank2016dimensions}. Therefore, the over-representation of educated respondents makes our findings more conservative, likely providing lower-bound estimates: if significant digital divides and modest literacy levels are evident even in this highly educated, digitally connected sample, they are likely more pronounced in the broader population.\footnote{Our literacy results align with the 2025 IPSOS monitor, which uses a representative sample of the Italian adult population (online sample, age 18-75: 40\% of their participants disagreed with the statement ``I have a good understanding of what artificial intelligence is" \citep{ipsos2025aimonitor}. This is consistent with our survey, where 43.3\% of respondents disagreed that they had a ``good knowledge of (AI) language technologies.''}

To test the robustness of our results, we performed sensitivity analyses using post-stratification weights based on nationally representative ISTAT data for gender, age, and geography. The findings remain substantively unchanged (results in Supplementary Section~\ref{appB:robustness}). Therefore, we retain all data to maximize statistical power. 

Our sample shows a higher proportion of GenAI users (80.4\%, $n=1533$, 80.1\% in the weighted sample) compared to non-users.  
We acknowledge that, as an opt-in survey, individuals who do not use the technology may have been less inclined to participate. 
Nevertheless,  non-users ($n=373$) remain sufficient for robust analysis of adoption trends.

\subsection{Measures}
\label{subsec:measures}

We operationalize key variables of interest.

\textbf{GenAI chatbot adoption.} 
We treat GenAI chatbot adoption as a binary variable based on participants' self-reported use of GenAI chatbots in the past 12 months. To facilitate understanding, we provided a multiple-choice list of specific chatbot models, such as ChatGPT, Claude, and Gemini (list available in the Supplementary Table~\ref{tab:definitions}, Section~\ref{appA:definitions}). We classified participants as \textit{users} if they selected at least one of these models and as \textit{non-users} if they did not select any.

\textbf{LT adoption and experience.} 
To compare GenAI chatbot adoption with the use of longstanding task-specific LTs, we collected data on the use of five applications: machine translation, voice assistants, assisted writing, speech transcription, and text-to-speech. Responses for each application were coded as binary variables (has/does not have experience). 
An index of prior LT experience was then constructed by summing participants' responses for each of the five applications, with the resulting score ranging from 0 (indicating no use of any application) to 5 (indicating use of all five applications).

\textbf{LT Training.} We asked respondents whether they had received 
any training on GenAI chatbots or AI language technologies, distinguishing between: formal training received at school or in the workplace,  self-directed learning (e.g., online tutorials), other AI-related 
coursework, or none. We dichotomized responses into a binary variable (\textit{Trained} $n=837$
and \textit{Untrained} $n=1069$).

\textbf{LT literacy.} 
We assessed self-reported literacy of GenAI and AI language technologies more broadly using six items on a 5-point Likert scale (ranging from strongly disagree to strongly agree). 
Drawing on  \citeauthor{long2020}, we focus on functional and critical dimensions of literacy as proxies for skills and competence.
The items included: (1) ``I have a good knowledge of language technologies,'' (2) ``I can recognize errors in language technologies,'' (3) ``I can distinguish automatically generated text from text written by humans,'' (4) ``I feel prepared to use language technologies,'' (5) ``I know the limitations of language technologies,'' and (6) ``I know the potential of language technologies.'' 
For inferential analyses (see Section \ref{subsec:analysis}), responses to these items were recoded into a mean unified score on a 0-1 linear scale, with higher values indicating greater self-assessed LT literacy.
To assess potential reporting bias, we validate this measure using regression analyses (full results in Supplementary Section~\ref{app:ssec:training_literacy_moderation_analysis}, Table \ref{tab:training_literacy} and \ref{tab:training_literacy_gender}). LT training is a significant positive predictor of self-assessed literacy among both GenAI chatbot users ($\beta = 0.13$, $p < .001$) and non-users ($\beta = 0.099$, $p < .001$), consistent with the expectation that training enhances literacy. Besides, the association of LT training with self-assessed literacy is moderated by gender: training yields significantly smaller gains in self-reported literacy for women than for men (Training $\times$ 
Woman: $\beta = -0.06$, $p = .002$), echoing evidence that women self-report comparatively lower digital skills  \citep{hargittaiwomen}. We thus retain both measures as complementary: training as a more factual proxy of acquired education, and self-assessed literacy as a subjective measure more closely related to perceived competence, which can nonetheless shape engagement.



\textbf{Usage intent type and frequency.} 
For GenAI users, we collected self-reported data on the frequency of GenAI chatbot use across six specific intents: Information Retrieval, Problem Solving, Learning, Content Creation, Leisure, and Creativity (see Table~\ref{tab:intent}). The response options were recoded to represent frequency on a scale of 0 (never), 1 (less than monthly), 2 (monthly), or 3 (weekly).
%
This measure aims to capture the quality and quantity of engagement modes. We anticipate distinguishing intents across two axes: 
active vs. passive and instrumental vs. recreational.  Content Creation, Learning, and Creativity 
map onto active, instrumental uses with attested productive and 
capacity-augmenting value in the literature \citep{noy2023experimental, 
brynjolfsson2025generative, abbes2024generative}. Problem Solving shares 
this active, instrumental orientation, but evidence of its benefit 
potential remains limited.\footnote{\citet{regenwetter2025generativeoptimizationperspectiveaienhanced} 
argues for the value of GenAI problem-solving capabilities, but 
restricts this claim to the engineering domain.} 
Leisure is associated with 
recreational, lower-capital engagement. 
While certain forms of purposive information-seeking have been associated with capital-enhancing value in prior 
internet use research \citep{hargittai2008digital}, here Information Retrieval approximates 
a more passive, generic search-like behavior---and one of inferior reliability 
given GenAI's documented limitations in providing factual information 
\citep{chen2024combating}. 
We treat this mapping as an exploratory heuristic: the rapid evolution of GenAI chatbots and the scarcity of research directly examining differential returns across usage types imply that definitive classifications remain premature. We return to this 
framing in the interpretation of our results. 

\subsection{Analysis and Methods}

\subsubsection{Statistical Analysis}
\label{subsec:analysis}
Our analytical approach combines descriptive and inferential methods. We use descriptive statistics---e.g., frequencies, means, standard deviations, and cross-tabulations---to characterize broad patterns across the sample, such as GenAI chatbots and LTs adoption rates, the distribution of LT training and literacy, and overall coverage of GenAI chatbot activities. Instead, we use inferential methods to explore divides in adoption and usage frequency across intents.


We use logistic regression to predict GenAI chatbot adoption (users vs. non-users) and estimate odds ratios, which provide a more intuitive measure of the likelihood of adopting GenAI based on the predictor variables. 

To identify intent-specific predictors of usage frequency, we run separate ordinary least squares (OLS) models for each of the six intents. While this is formally an ordered measure (0-3),  
it can be reasonably interpreted as approximating a continuous frequency of use. In fact, ordered logistic and OLS regressions yield virtually identical results in terms of sign, magnitude, and significance of the coefficients (coefficient correlation across all six models is $r = 0.993$, $p < .001$, significance sign agreement~$=~98.7\%$). For parsimony and ease of interpretation, we therefore report OLS estimates, which include LT experience, training, and literacy as independent variables. Full results across models are reported in Supplementary Section~\ref{app:ssec:complementary_results}, Tables~\ref{table:lit-stats} and \ref{tab:ols_intent} for completeness.

All models incorporate sociodemographic variables as controls, excluding participants who identified as other than man or woman due to small size ($n = 56$, see Table \ref{tab:sample_description}). For adoption divides, we also include LT experience as an independent variable to account for technology familiarity. For OLS usage models across intents, we additionally include LT training and literacy to account for digital competence.  

We verified that the predictor variables do not exhibit problematic multicollinearity using Variance Inflation Factors \citep[VIF;][]{o2007caution}. All VIF values are below 1.3, well below conventional thresholds, indicating no significant multicollinearity among predictor variables (full results reported in Supplementary Table~\ref{tab:VIF}, Section~\ref{app:ssec:multicollinearity}).

\subsubsection{Free-Form Response Analysis}

Our survey included optional open-ended questions to capture qualitative insights into participants' perceptions of the technology's usability, experience with potential errors, and preferences for use (i..e., language of the interaction). We include in our analyses three questions that received at least 300 responses ($\sim$20\% of the users' response pool). 

Following standard textual modeling practices \citep[e.g.,][]{Alipour2023CrossplatformSD,Bhandarkar2025InvestigatingTP}, we search for and group similar answers into clusters, and then describe each cluster's topic to infer trends and insights.
Given all open-text responses of a question, we use BERTopic \citep{grootendorst2022bertopic}, an established topic modeling tool. The software uses embedding models to transform text into numerical representations, clustering and dimensionality reduction algorithms, and LLMs to generate textual descriptions of the discovered topics. 
We report in the Supplementary Section~\ref{ssec:app:bertopic} more details about BERTopic, as well as the complete information on the preprocessing applied to texts, our choices of model parameters, and steps taken to verify the extracted topics. 

\section{Results}\label{sec:Results}
Our results are organized as follows: Section \ref{subsec:1-results} explores GenAI and LTs adoption rates and first-order divides. Section \ref{subsec:2-results} examines participants' literacy and their awareness of potential benefits, risks, and social impact.
Section \ref{subsec:3-results} focuses on second-order divides in usage and examines the range, context, and purpose of activities in which GenAI chatbot users engage.  

\subsection{GenAI and Language Technologies Adoption}\label{subsec:1-results}
As a first layer of analysis, we examine the extent and drivers of GenAI chatbot adoption by comparing it with other  LTs. Specifically, we investigate which demographic and socioeconomic factors are associated with higher or lower likelihood of adoption, as well as prior LT experience. We complement this analysis with qualitative insights into the reasons behind (un)willingness to adopt.

\begin{figure}[htp]
    \centering
\includegraphics[width=1\linewidth]{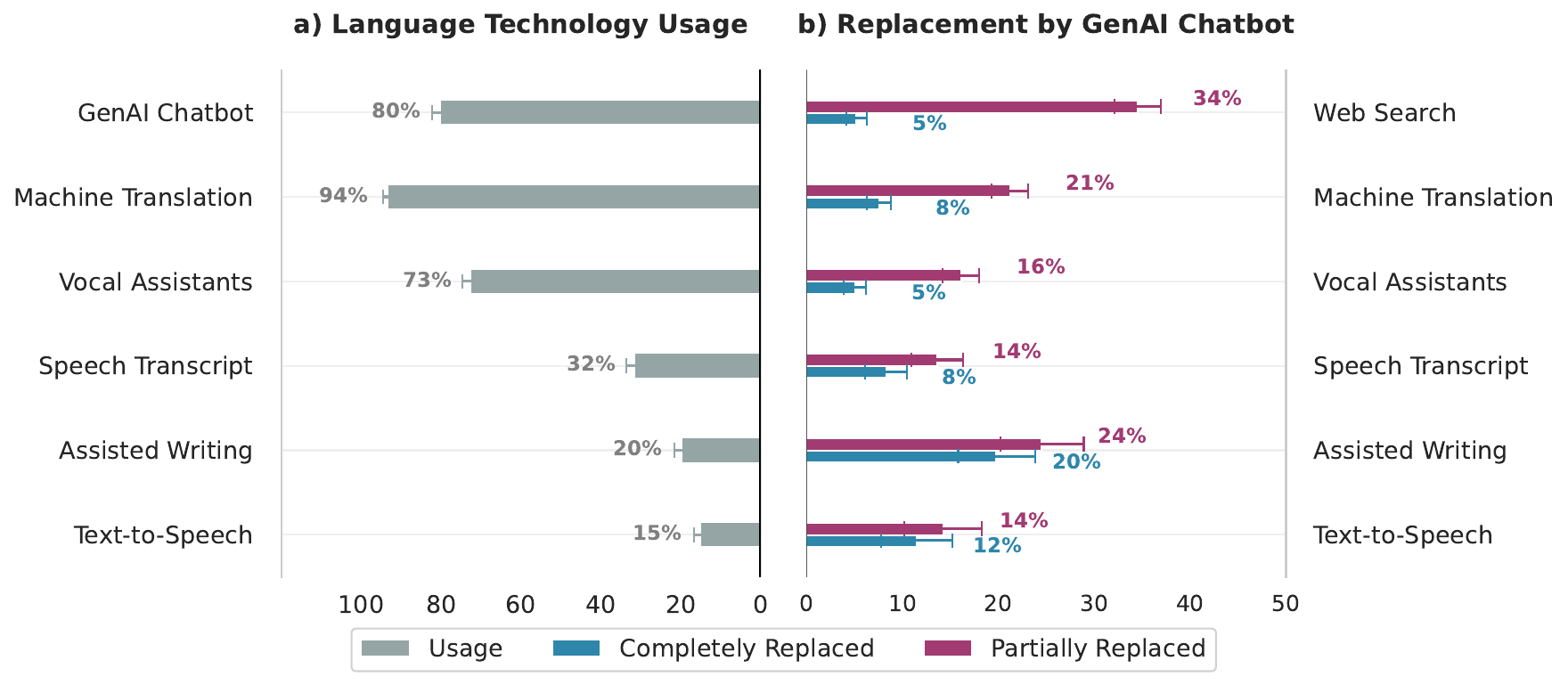}
    \caption{\textbf{Language technology adoption and replacement}.
     Panel.\textbf{a} (\textit{left}) shows the ratio of participants  who reported having used each language technology, including GenAI chatbots. Panel.\textbf{b} (\textit{right}) shows, among users of each technology, the ratio of users reporting that GenAI chatbots have completely or partially replaced their use of that technology. For ``Web search'', ratios are calculated among GenAI chatbot users. Error bars represent 95\% confidence intervals.}
    \label{fig:usage_shift_breakdown}
\end{figure}

\paragraph{Adoption Rates and Drivers}
Figure~\ref{fig:usage_shift_breakdown}.a presents adoption rates across six LTs, offering a comparative view of the extent of GenAI chatbot diffusion. GenAI chatbots adoption (80\%) follows the near-universal usage of machine translation (94\%), but surpasses established technologies such as voice assistants (72\%) like Siri and Alexa, which have been popular for years. By contrast, speech transcript and text-to-speech remain considerably less prevalent, suggesting that longevity alone does not guarantee uptake. Besides, our data show that GenAI chatbots centralize activities from other technologies. 
When GenAI users were asked whether the onset of GenAI chatbots had affected how and if they used other digital technologies, 35\% reported a definitive shift, while an additional 18\% were unsure. Technology-specific patterns are visible in Figure~\ref{fig:usage_shift_breakdown}.b, with assisted writing facing the most substantial replacement (24\% partial, 20\% complete) and web search experiencing meaningful displacement (34\% partial, 5\% complete), indicating that users complement their information diet bypassing traditional online queries.\footnote{We note that AI-augmented search features (e.g., Google's AI Overviews, launched in Italy in March 2024) may also contribute to web search displacement~\citep{giuffrida2025google}. To mitigate this ambiguity, our survey specifically targets GenAI users only, respondents who access GenAI chatbots via conversational interfaces.}

To understand GenAI chatbot diffusion, we asked participants who reported replacing other technologies  ($n=624$) to select reasons from six multiple-choice options (i.e., convenience, flexibility \& personalization, accuracy, reliability, transparency, besides ``other''). The results reveal that the shift stems primarily from practical advantages: convenience (38\%) and flexibility/personalization (30\%) dominate users' motivations, while accuracy (19\%), reliability (8\%), and transparency (2\%) play secondary roles. 

\begin{table*}[htp!]
\centering
\label{tab:logit_adoption_m1m2}
\rowcolors{2}{white}{gray!15}
\small
\begin{tabular}{lcc}
\toprule
 & \textbf{Model 1} & \textbf{Model 2} \\
\midrule
Gender: Woman & \textbf{0.404$^{***}$} & \textbf{0.436}$^{***}$ \\
  & \small (0.053) & \small (0.059) \\
Age: 35--54 & \textbf{0.389$^{***}$} & \textbf{0.423$^{***}$} \\
  & \small (0.082) & \small (0.090) \\
Age: 55--64 & \textbf{0.260$^{***}$} & \textbf{0.319$^{***}$} \\
  & \small (0.055) & \small (0.070) \\
Age: 65+ & \textbf{0.108$^{***}$} & \textbf{0.149$^{***}$} \\
  & \small (0.023) & \small (0.032) \\
Geography: Centre & 1.068 & 1.057 \\
  & \small (0.190) & \small (0.193) \\
Geography: South \& Islands & 0.808 & 0.834 \\
  & \small (0.138) & \small (0.146) \\
Geography: Abroad & 1.105 & 1.139 \\
  & \small (0.365) & \small (0.383) \\
Income: Mid & 1.309 & 1.275 \\
  & \small (0.222) & \small (0.222) \\
Income: Higher & \textbf{1.728$^{**}$} & \textbf{1.710$^{**}$} \\
  & \small (0.350) & \small (0.356) \\
Education: Graduates & \textbf{1.720$^{***}$} & \textbf{1.632$^{***}$} \\
  & \small (0.234) & \small (0.229) \\
LT Experience & --- & \textbf{1.859$^{***}$} \\
  & \small  & \small (0.133) \\
  \midrule 
Observations & 1873 & 1873 \\ 
Pseudo $R^2$ & 0.128 & 0.175 \\
$\chi^2$ & 235.90 & 322.22 \\
$\Delta\chi^2$ (M2 vs M1) & --- & 86.32$^{***}$ \\
\bottomrule
\end{tabular}
\caption{\textbf{GenAI Chatbot Adoption (Odds Ratios)}. Logistic regression results predicting GenAI chatbot adoption. 
Model 1 includes sociodemographic predictors only; Model 2 adds prior experience (LT Experience). Odds ratios are reported with standard errors in parentheses. Statistically significant results are in bold: $^{*}p<0.05$; $^{**}p<0.01$; $^{***}p<0.001$. Participants who identified as neither man or women are excluded due to small size. References: Men, 18-24, North, Lower Income, Non-graduates.}
\label{tab:combined_logit}
\end{table*}

\begin{figure}[htp!]
    \centering
    \includegraphics[width=0.5\linewidth]{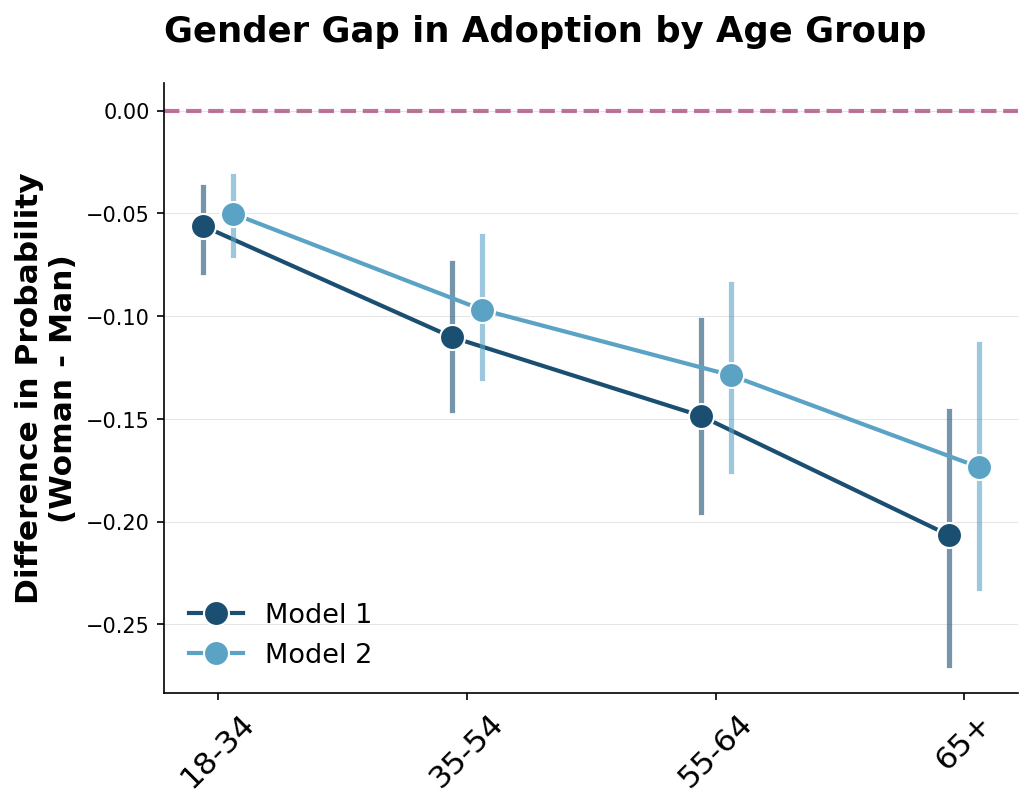}
\caption{\textbf{Gender gaps in GenAI chatbot adoption}. Average marginal effects of gender (Woman - Man) across age groups with 95\% bootstrap confidence, as estimated by logistic regression. Negative values indicate lower predicted adoption by women. Model 1 controls for sociodemographic correlates, while Model 2 also includes prior LT experience.}
    \label{fig:ame_gender}
\end{figure}

\paragraph{Adoption Divides}
Next, we focus on the social structuring of GenAI adoption by modelling adoption likelihood as a function of sociodemographic characteristics and prior experience with other LTs.  Table~\ref{tab:combined_logit} presents odds ratios from two logistic regressions predicting adoption likelihood. Model 1 relies solely on sociodemographic variables, whereas Model 2 also includes prior LT experience to examine the effect of technology familiarity. 
Model 1 reveals clear sociodemographic fault lines. 
Consistent with our expectations, 
odds of adoption decline with increasing age, and women show 60\% lower odds of adoption compared to men (OR = 0.40). 
Graduate education (OR = 1.72) and higher income (OR = 1.73) are associated with greater adoption, while geography plays no significant role.
In Model 2, the effects of economic and sociodemographic factors are attenuated but remain significant. Prior experience, emerges as the strongest positive predictor (OR = 1.86), nearly doubling the odds of adoption per unit increase and improving model fit. 
Focusing on negative correlates of adoption, Figure~\ref{fig:ame_gender} plots the average marginal effects of gender on adoption probability by age group for both models. The gender gap widens markedly with age: women's adoption probability is 5.0 percentage points lower than men's among 18--34 year-olds, rising to 17.3 points among those aged 65+ (Model 2). This gradient is robust across both model specifications, showing that gender and age jointly compound into a persistent adoption divide, and that gender gaps are smaller among younger cohorts.

\begin{figure}[htp!]
    \centering
    \includegraphics[width=0.5\linewidth]{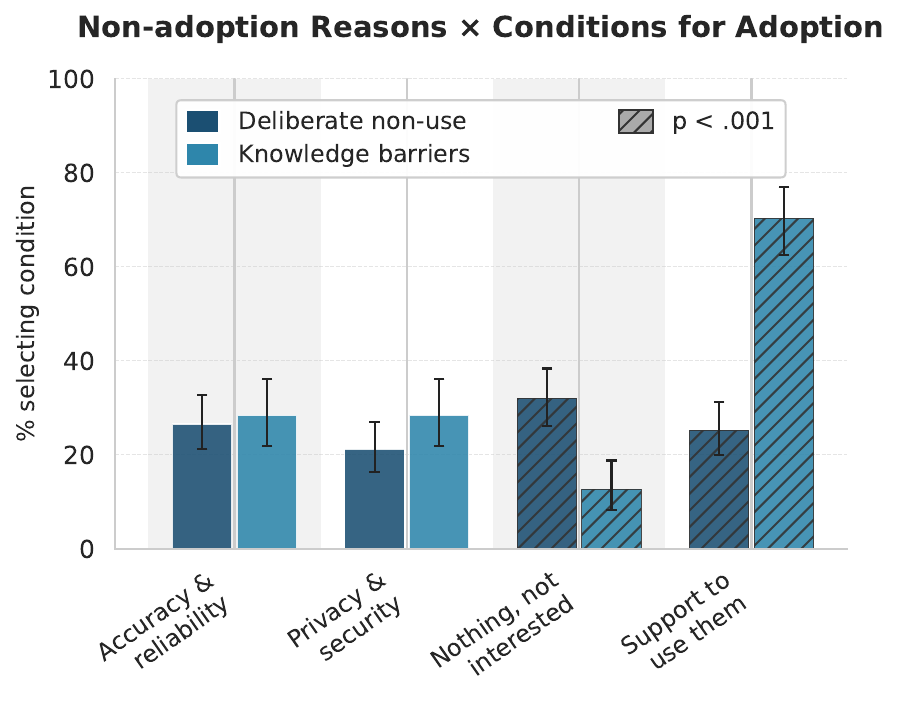}
    \caption{\textbf{Non-adoption reasons and conditions for future adoption among non-users.} Two non-user groups are compared: deliberate non-users, who reported not adopting GenAI chatbots by preference or perceived lack of utility, and non-users who face knowledge or skills barriers. Shaded bars represents the proportion of respondents selecting a given condition for future adoption. Error bars represent 95\% confidence interval. Statistical significance is measured as per-condition chi-square tests of independence.}
    \label{fig:non_user_segment}
\end{figure}


\paragraph{Non-adoption Motivations}
To bring into focus the non-adoption segment, we asked 
non-users  about their reasons for not using GenAI chatbots and 
what conditions might facilitate future adoption. We identify two substantially 
different groups: those facing knowledge and skills barriers---i.e., unaware 
of chatbot existence or unable to use them (40.5\%, $n =151$)---and 
deliberate non-users, who reported not adopting them by preference or perceived 
lack of utility (59.5\%, $n =222$). Notably (see Figure~\ref{fig:non_user_segment}), 
these groups share a need for privacy and reliability guarantees---but diverge substantially in 
their conditions for future adoption. Non-users with knowledge barriers concentrate around support in understanding and using  
GenAI chatbots (70.2\%), consistent with a usability deficit that  
intervention could plausibly address. Voluntary non-users, instead, are more likely to express complete disinterest in future use 
(31.9\% vs.\ 12.6\%).



\subsection{Language Technology Literacy and Impact}\label{subsec:2-results}

Whereas the previous section focused on adoption as first site of inequality, this section turns to technology literacy and awareness of impact as a further axis of differentiation. We examine participants' proficiency with LTs and their knowledge of associated risks, benefits, and broader social implications. We draw on prior training exposure and self-reported literacy across multiple dimensions (Sec.~\ref{subsec:measures}).

\begin{figure}[htp!]
    \centering
    \includegraphics[width=1\linewidth]{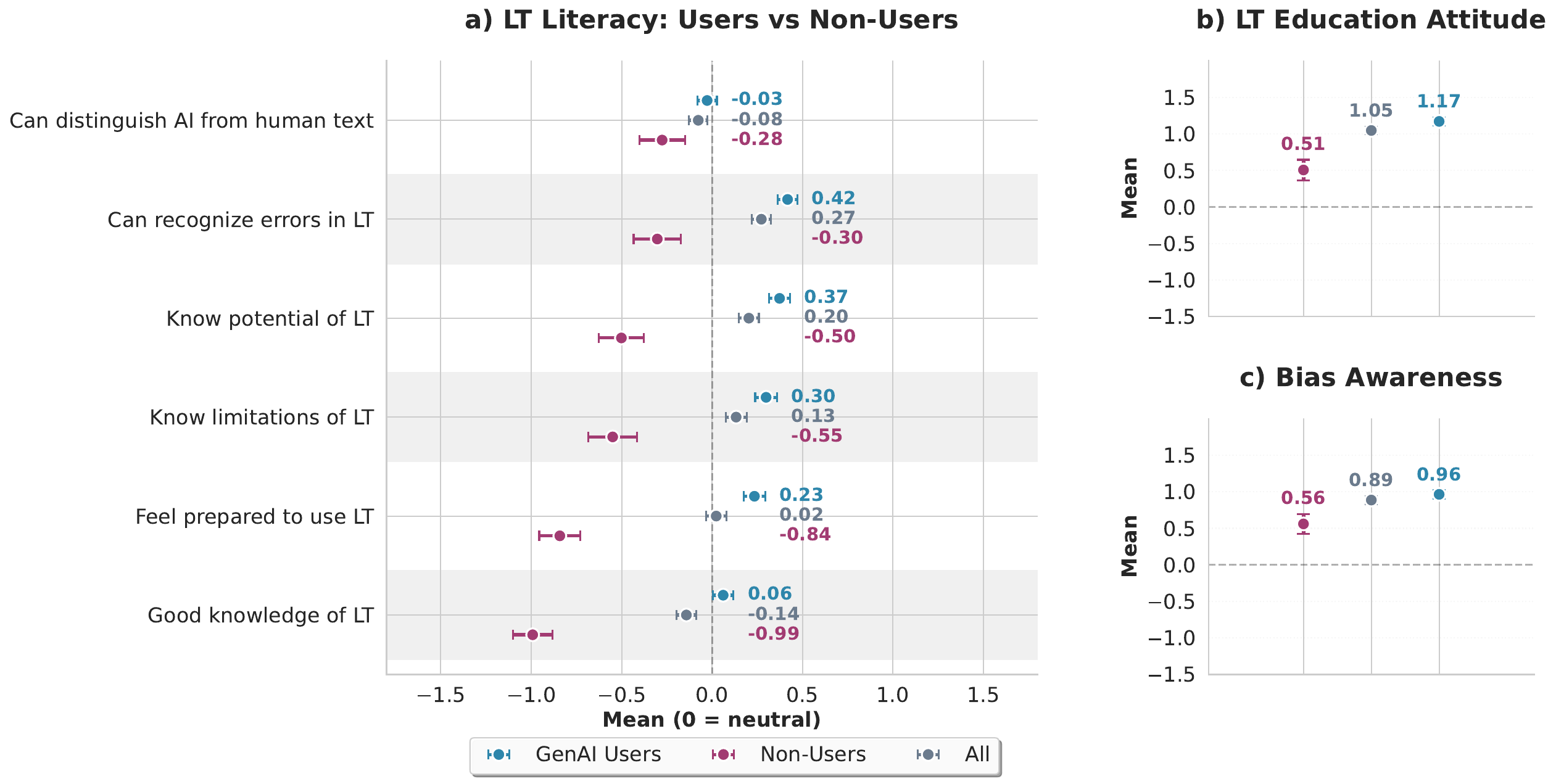}
    \caption{\textbf{Mean responses on LT literacy, attitudes toward LT education, and bias awareness among GenAI users and non-users}.  Responses are measured on a five-point scale, plotted as $-2$ (strongly disagree) to $+2$ (strongly agree), with $0$ representing a neutral stance. 
    We use a restricted range ($-1.5$---$1.5$)  for better visualization. 
    Panel.\textbf{a} (\textit{left}) shows six LT literacy items, with dots representing the group means. Panel.\textbf{b} and Panel.\textbf{c} depict, respectively, the believed importance of establishing LT education in schools and higher education (top) and awareness that LT can reproduce biases (bottom). All differences between users and non-users are statistically significant according to independent samples t-tests ($p < 0.001$). Effect sizes (Cohen's $d$) for LT literacy items are consistently large (range: $d = 0.64$--$0.93$), with the exception of distinguishing AI from human text ($d = 0.22$). Full descriptive statistics and effect sizes are reported in Supplementary Section~\ref{app:ssec:complementary_results}}
    \label{fig:lit}
\end{figure}


\paragraph{Education and Literacy Levels}

We find limited prior education in LT among respondents:
most participants reported receiving no formal training or informal 
on LTs, nor any other AI technology. This knowledge gap is particularly pronounced among non-users, with 76.2\% reporting no training at all compared to 51.2\% of users.
%
Self-assessed literacy levels mirror this training deficit. As shown in Figure~\ref{fig:lit}.a, mean scores on a five-point scale---
with 0 representing a neutral stance---reveal modest self-perceived competence, even among chatbot users, though scores are even lower for non-users. Participants reported greatest difficulty with theoretical knowledge (``good knowledge of LT'': $M = -0.99$, non-users; $M = 0.06$, users; $d=0.93$, $p < .001$) compared to practical applications (``feel prepared to use LT'': $M = 0.23$, users; $M = -0.84$, non-users; $d=0.92$, $p < .001$). Notably, both groups expressed high uncertainty recognizing AI-generated text (users: $M = -0.03$; non-users: $M = -0.28$; $d=0.22$, $p < .001$).
Literacy levels remain low even among chatbot users: disaggregated results by age (Table~\ref{tab:user_lit}) reveal a generational gap, with younger cohorts reporting higher competence compared to older groups.


\begin{table}[]
\begin{tabular}{lcccc}
\hline
\textbf{LT Literacy (GenAI Users)} & \textbf{18-34}               & \textbf{35-54}               & \textbf{55-64}                & \textbf{65+}                  \\ \hline
Good knowledge of LT               & \cellcolor[HTML]{D2E4F4}0.18 & \cellcolor[HTML]{D5E6F5}0.14 & \cellcolor[HTML]{D9E8F6}0.08  & \cellcolor[HTML]{FAFCFE}-0.42 \\
Feel prepared to use LT            & \cellcolor[HTML]{A4C5DA}0.34 & \cellcolor[HTML]{A8C8DC}0.33 & \cellcolor[HTML]{C0D8EA}0.26  & \cellcolor[HTML]{EFF5FB}-0.25 \\
Know limitations of LT             & \cellcolor[HTML]{78A7C0}0.47 & \cellcolor[HTML]{93B9D0}0.39 & \cellcolor[HTML]{C3DAEC}0.25  & \cellcolor[HTML]{EBF3FA}-0.19 \\
Know potential of LT               & \cellcolor[HTML]{669BB6}0.52 & \cellcolor[HTML]{8CB5CC}0.41 & \cellcolor[HTML]{A1C3D8}0.35  & \cellcolor[HTML]{DEEBF7}0.00  \\
Can recognize errors in LT         & \cellcolor[HTML]{5F96B2}0.54 & \cellcolor[HTML]{5891AD}0.56 & \cellcolor[HTML]{93B9D0}0.39  & \cellcolor[HTML]{E5EFF9}-0.10  \\
Can distinguish AI from human text & \cellcolor[HTML]{D7E7F5}0.11 & \cellcolor[HTML]{DDEAF7}0.02 & \cellcolor[HTML]{E2EEF8}-0.06 & \cellcolor[HTML]{F6FAFD}-0.36 \\ \hline
\textbf{Average}                   & \cellcolor[HTML]{AFCCE0}0.36 & \cellcolor[HTML]{C5DBED}0.31 & \cellcolor[HTML]{D4E5F5}0.21  & \cellcolor[HTML]{EEF5FB}-0.22 \\ \hline
\end{tabular}
\caption{\textbf{LT literacy scores by Age Group for GenAI users}. Responses are measured on a five-point scale ranging from $-2$ (strongly disagree) to $+2$ (strongly agree), with $0$ representing a neutral response. 
}
\label{tab:user_lit}
\end{table}


Given the  moderate levels of LT literacy and previous education reported, Figure \ref{fig:lit}.b presents mean scores on endorsing the integration of LT education into formal school and university curricula.
GenAI users express stronger support than non-users
($M = 1.17$ vs. $0.51$; $d=0.58$, $p < .001$), likely reflecting the segment of non-users who have deliberately opted out of GenAI adoption.


\paragraph{Social Bias and Hallucination Awareness}

Despite limited overall literacy end education received, participants demonstrate comparatively higher awareness that LTs can reproduce biases and stereotypes (Figure \ref{fig:lit}.c: users $M = 0.96$, non-users $M = 0.56$; $d=0.34$, $p < .001$ ). 
Also, overall bias awareness (All $M = 0.89$) scores higher than the ability to spot errors (All $M = 0.27$), potentially signaling a gap between awareness and actual ability to recognize such issues. 
This gap is further evidenced by GenAI users' reported experiences: when asked if they had ever encountered  errors or biased examples in the output of GenAI chatbots, 39.5\% of GenAI users confirmed encountering them, whereas 24.3\% were unsure.

In a free-form question, we asked participants to expand on their experienced errors and stereotyped outputs. Among those who commented ($n=334$), the most commonly reported issues centered on factual inaccuracies and fabrications. 
The largest cluster ($n=100$) describes experiences with hallucinations---instances in which GenAI chatbots ``present false information as facts.'' Issues frequently concern ontological knowledge about Italian culture, e.g., historical events or bibliographic details. Crucially, some ($n=6$) report issues in high-stakes contexts, e.g., legal and fiscal requests about the Italian system, with 
GenAI generating ``references to non-existent legal articles,'' ``fabricating supporting precedents,'' or offering information that ``contradicts established Italian law.''
Several users ($n=55$) documented pervasive gender and racial bias in AI outputs.
In image generation, users reported managerial roles depicted as ``handsome middle-aged men,'' teachers as young, blond, ``Barbie-like'' women figures, and caregivers as inherently feminine.
Requests for ``images of poor people consistently generated people of color,'' while successful people yielded white individuals.
Linguistic biases were also prevalent ($n=24$), with users reporting a systematic defaulting to masculine forms in Italian, even when gender-neutral options existed in the source language---e.g., ``translating a doctor and a nurse'' consistently produced ``un medico e un'infermiera,'' (masculine and feminine forms in Italian, respectively). Additionally, some users ($n=11$) identified cultural homogenization, describing responses as ``anglocentric'' and noting the GenAI's tendency toward excessively polite outputs that avoid genuine depth or critical perspectives. 

\subsection{GenAI Usage }
\label{subsec:3-results}

Having examined adoption rates, divides, and literacy levels, 
we turn to GenAI users and examine how they engage with the technology in practice. We start by broadly characterizing the purpose of usage and activities across professional and personal contexts. Then, we focus on differentials across interaction types, diversity, and, finally, frequency of usage across six different intents. 


\begin{figure}[t]
    \centering
    \includegraphics[width=1\linewidth]{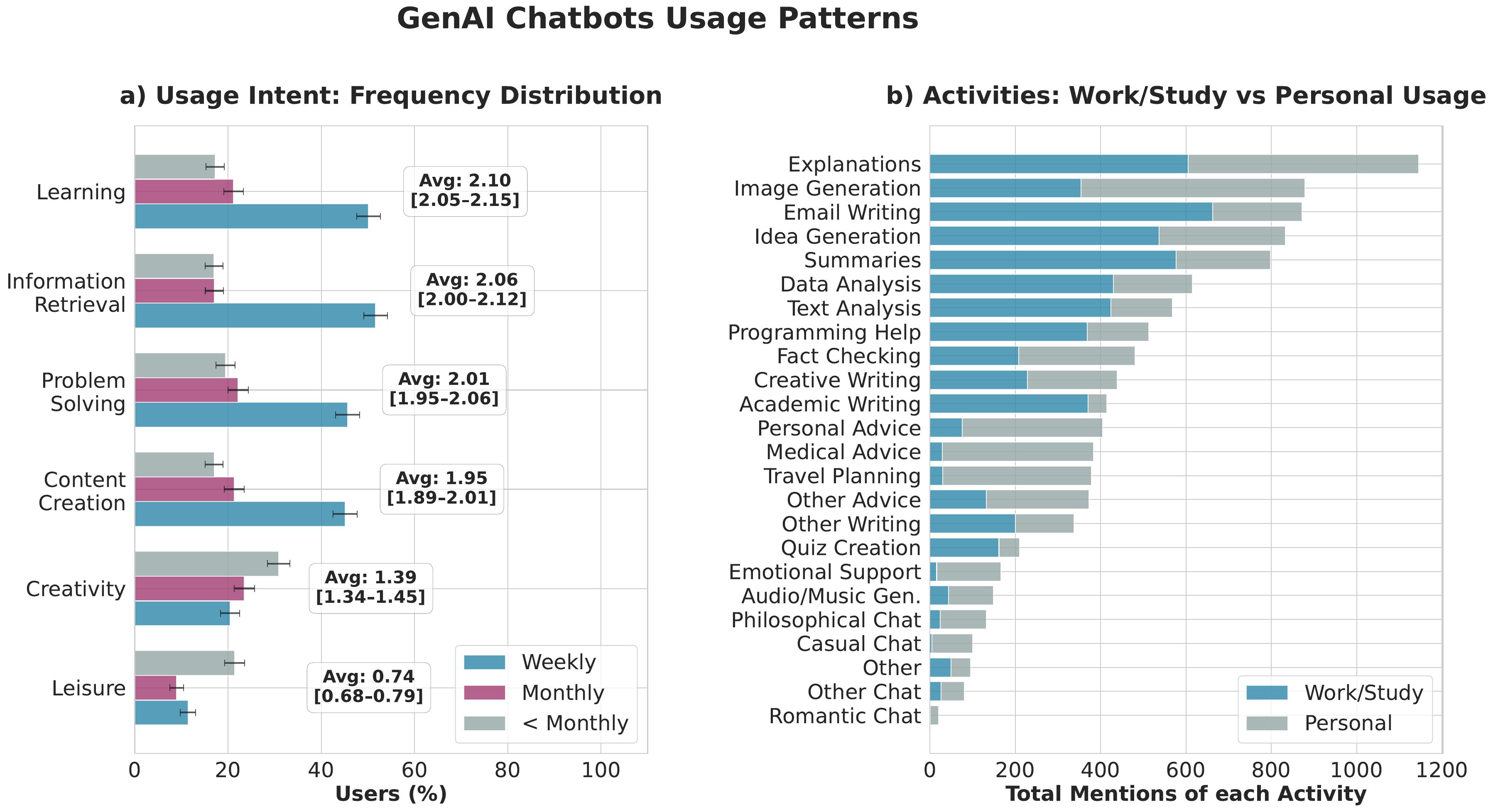}
    \caption{ 
    \textbf{Usage patterns for GenAI chatbots}. Panel\textsc{\textbf{.a}} displays frequency distributions across six usage intents with annotated mean frequency scores (0 = never, 1 = less than monthly, 2 = monthly, 3 = weekly) and 95\% confidence intervals  Panel.\textbf{b} shows the total count of activities performed across all users, distinguishing the number of times each activity was indicated for work/study ($n=5562$) versus personal contexts ($n=4827$). }
    \label{fig:genai_patterns}
\end{figure}

\paragraph{GenAI Chatbots as Multipurpose tools}

Broad descriptive trends on the multipurpose nature of GenAI chatbots are given in Figure \ref{fig:genai_patterns}.
Figure \ref{fig:genai_patterns}.a shows GenAI chatbot usage frequency across six intents. \textit{Learning} and \textit{Information Retrieval} lead with the highest average frequencies (2.1 and 2.06 on a 0--3 scale), with most users engaging weekly (51\% and 52\%). \textit{Problem Solving} and \textit{Content Creation} follow closely, but with lower weekly usage shares (46\% and 45\%). \textit{Leisure} shows the lowest average frequency (0.74), but with a 20\% of respondents that engage with GenAI chatbots for leisure regularly (monthly or more). 

Figure \ref{fig:genai_patterns}.b provides complementary, granular insight into specific activities performed with GenAI chatbots.
\textit{Explanations} emerge as the most common activity (selected by 52.1\% of users at least once in either work/study or personal contexts), followed by \textit{Image Generation} (46.2\%) and \textit{Email Writing} (46.1\%). While instrumental activities dominate, the long tail contains activities such as \textit{Emotional Support} (10\%) and \textit{Medical} (23.9\%) or \textit{Personal Advice} (23.1\%), though less frequent, represent substantial and high-risk applications.\footnote{Activity types notably vary across age groups: younger users predictably dominate technical tasks such as data analysis, text analysis, and programming help. Older users (65+) stand out for seeking medical advice---likely driven by healthcare needs---and fact-checking. Additional qualitative results on this activity breakdown are reported in Supplementary Figure~\ref{fig:activities_age}, Section~\ref{app:ssec:complementary_results}.}  Overall, activity mentions show relatively balanced distribution between work/study (52\%) and personal (48\%,) context of usage, though this varies substantially by activity type. 

Regarding the language used to interact with GenAI chatbots,  while almost all users report using Italian (95\%), a substantial portion also relies on English (52\%). This figure is, however, stratified: 56.5\% of graduate users rely on English versus 38.2\% of non-graduates.
%
%
Our clustering analysis of free-form responses exposes the main users' attitudes. Among those who left an open comment about the \textbf{preferred language for interactions} ($n=529$), we identified several driving factors.
Most commonly ($n=207$), it is a contextual choice. Frequently, users prefer English for work and Italian for personal use.
Moreover, users associated queries in English with higher reliability in model outputs, especially for image generation.
A second trend ($n=95$) centers on a strong preference for Italian because it is the respondents' native language. Such explicit preference co-occurs with mentions to ``convenience,'' ``ease of use,'' ``practicality,'' and ``simplicity.'' 
In smaller clusters, language choice is explicitly related to 
i) the desired output language ($n=45$)---i.e., a pragmatic reason rather than a preference---
ii) task-dependent conditions ($n=32$), including the ``topic discussed,'' ``target audience'' or even their ``mood;'' 
iii) the superior availability of technical information in English ($n=27$) (e.g., for programming), and 
iv) educational purposes for themselves ($n=26$), e.g., to practice English. 
A final group ($n=26$)  uses GenAI chatbots almost exclusively in Italian due to a lack of competence in other languages, preventing them from interacting as they would like to. 
Thus, language selection emerges as a deliberate, context-dependent choice, balancing the convenience of native Italian with the perceived reliability and technical utility of English.

\paragraph{Activity Diversity and Contexts of Use}

We begin to characterize potential differences in usage patterns across groups by examining the diversity of specific activities users engage in, as well as the proportion of work/study vs. personal usage contexts. 

\begin{figure}[htp!]
    \centering
    \includegraphics[width=1\linewidth]{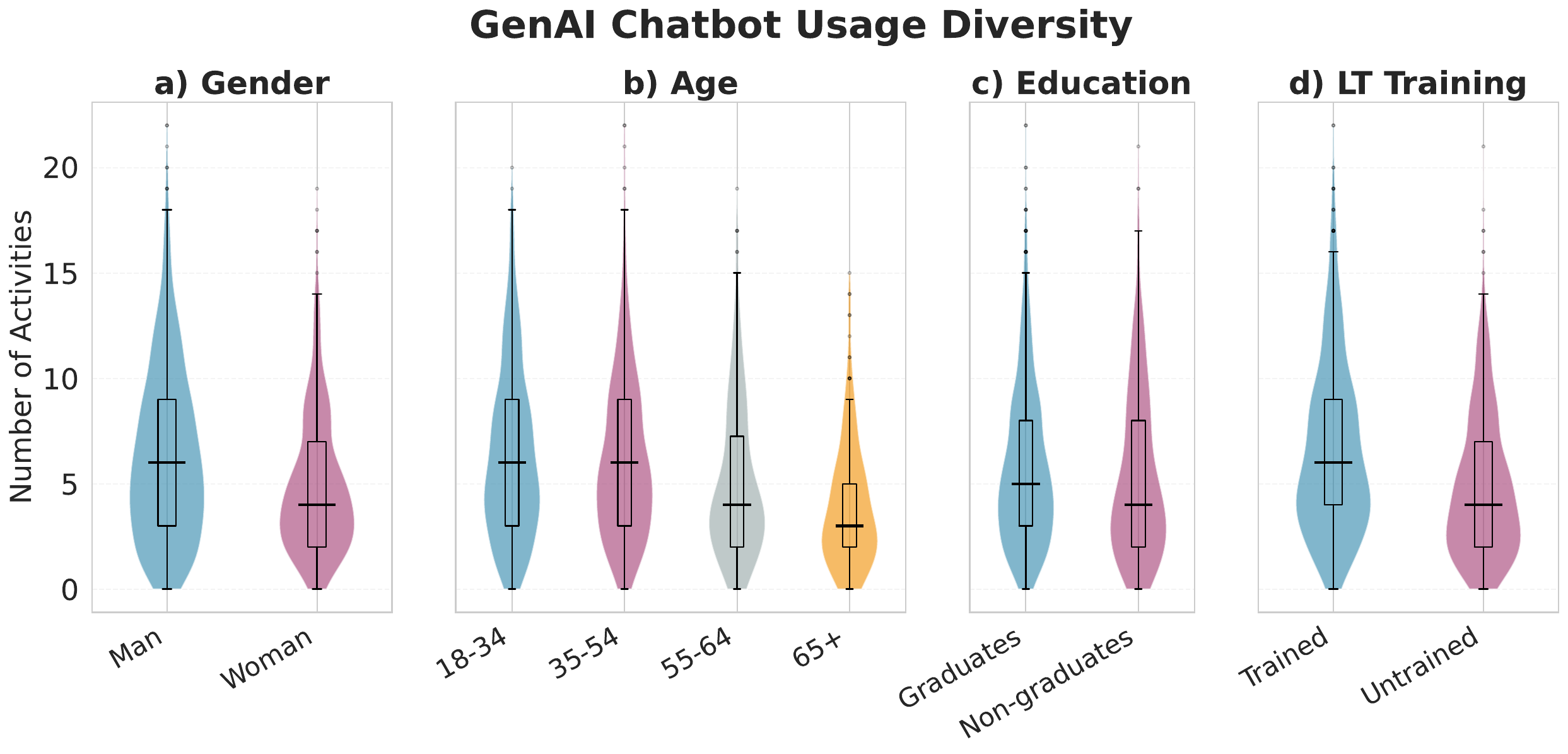}
    \caption{\textbf{GenAI chatbot Usage Diversity.} Distribution of the number of distinct activities performed by chatbot users across demographic groups. The violin shape shows the full data distribution. Group differences were statistically significant for gender ($H = 49.02$, $p < .001$), age ($H = 89.60$, $p < .001$), and LT training ($H = 74.31$, $p < .001$), as assessed by Kruskal-Wallis tests. Differences in Education are small ($H = 6.97$, $p < .01$)}.
    \label{fig:usage_diversity}
\end{figure}

GenAI Chatbot usage diversity---measured as the number of distinct activity types 
for which users reported using GenAI chatbots---is skewed in the sample (on average, out of 24 different activities, users engage in 5.73, median $=$ 5). This number significantly varies across sociodemographics (Figure \ref{fig:usage_diversity}): men report broader median usage compared to women (6 vs. 4), and diversity declines from 18-54 to 55--64 and 65+ (median = 6, 4 and 3, respectively). Also, 
trained users exhibit 
 broader median diversity than untrained users (6 vs.\ 4).


\begin{figure}[hpt!]
    \centering
    \includegraphics[width=1\linewidth]{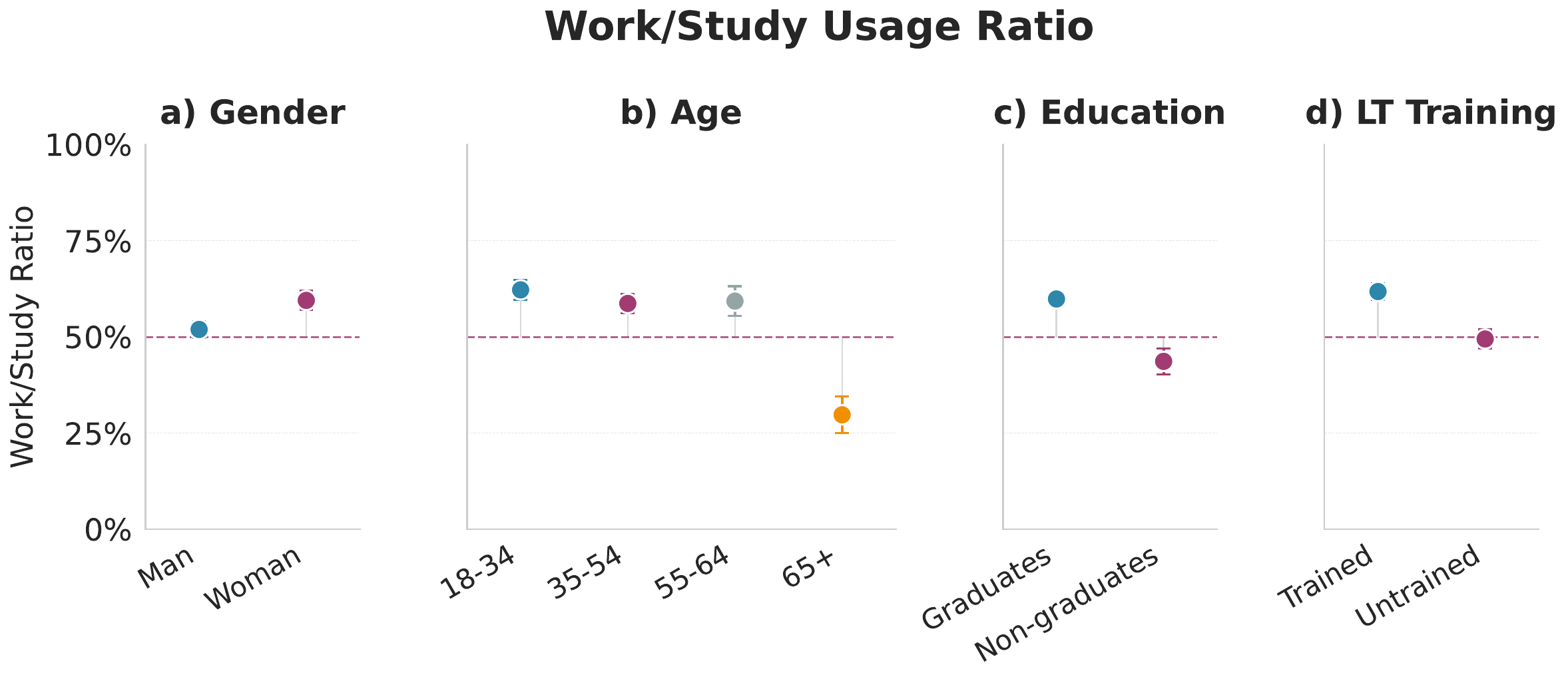}
    \caption{\textbf{Work/Study Usage Ratio.} Mean work/study usage ratio across demographic groups, where values above 50\% indicate predominant work-oriented use. Dots represent group means with 95\% confidence intervals. The dashed line marks the 50\% threshold. Group differences are statistically significant for gender ($H = 23.18$, $p < .001$), age ($H = 132.48$, $p < .001$), education ($H = 67.34$, $p < .001$), and LT training ($H = 48.01$, $p < .001$), as assessed by Kruskal-Wallis tests.}
    \label{fig:work_study_plot}
\end{figure}

Interestingly, results on the work/study ratio---defined as the share of activities 
devoted to work/study for each user, averaged within each group---provide a complementary picture (Figure \ref{fig:work_study_plot}). Those aged 
65+ showed markedly lower work-study dominant usage (30\%) compared to younger groups, in line with their status. 
In contrast, 
graduates and users with prior LT training direct a greater share of their 
use toward work and study than non-graduates (60\% vs. 44\%) and untrained users (62\% vs 50\%). The most compelling finding concerns gender: despite narrower usage diversity, 
women showed a \textit{higher} work/study ratio than men (60\%
vs. 52\%), suggesting their activities are more purposefully directed 
toward non-recreational contexts.

Regarding LT literacy, its relationship appears less clear-cut: it exhibits weak to moderate positive correlations with both usage diversity ($r = 0.29$, $p < .001$) and work/study orientation ($r = 0.17$, $p < .001$).

\paragraph{GenAI Usage Frequency by Intent}

\begin{figure}[hpt!]
    \centering
    \includegraphics[width=1\linewidth]{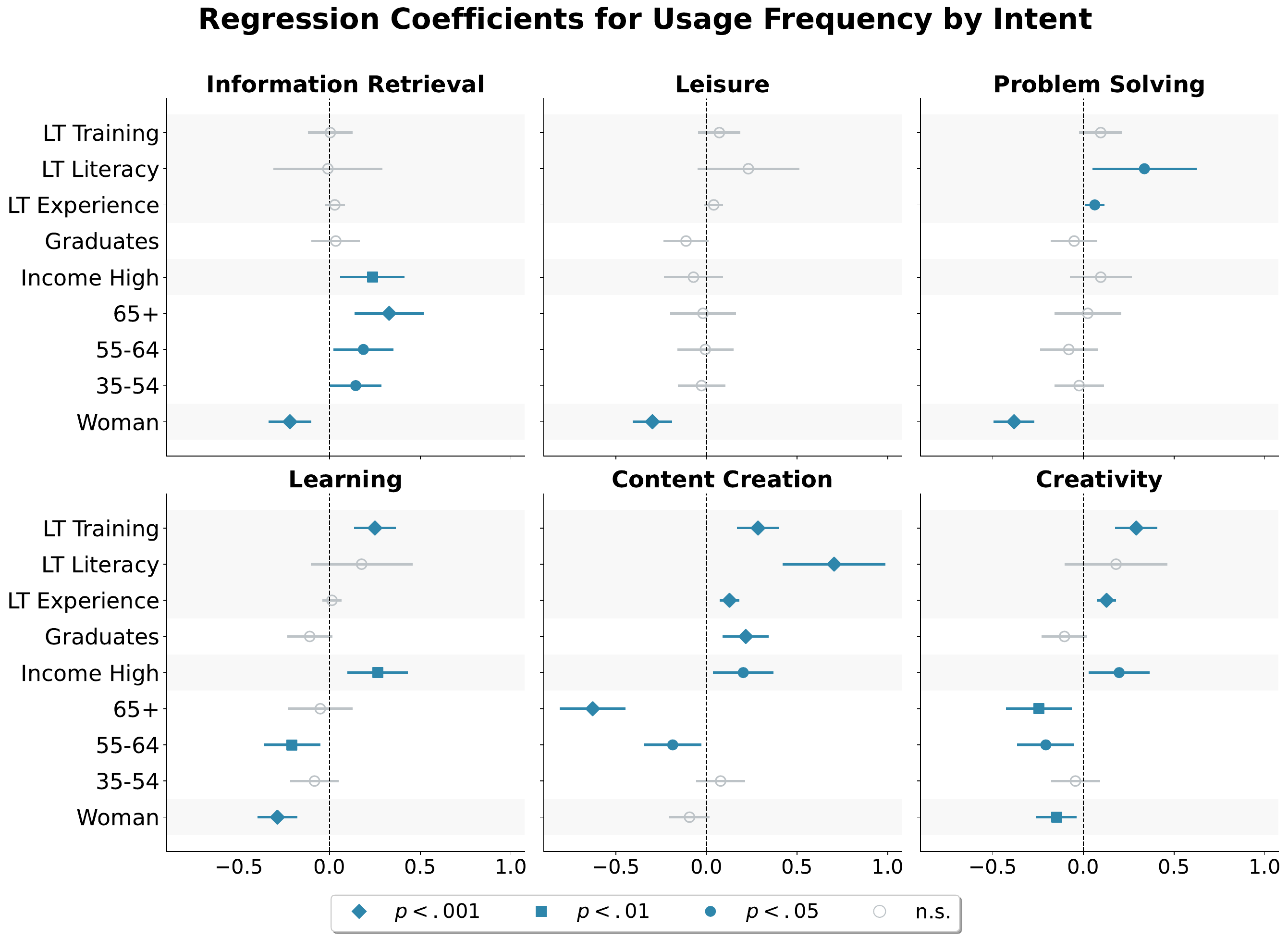}
    \caption{\textbf{Predictors of GenAI chatbot usage frequency by intent.} OLS regression coefficients with 95\% confidence intervals for six separate models predicting usage frequency across different intents among GenAI users. Models control for all sociodemographics, References: Men, 18-24, North, Lower Income, Non-graduates. Geography is excluded for ease of visualization. 
    Markers report statistical significance. For completeness, full coefficients are reported in Supplementary Table \ref{tab:ols_intent}, Section \ref{app_subsec:usage_frequency}.}
    \label{fig:ols_intents}
\end{figure}

To conclude, we examine the sociodemographic correlates of differences in usage type and frequency, as well as the effects of prior familiarity with LTs, training, and literacy levels. Figure 
\ref{fig:ols_intents} shows the results of six separate OLS regression models, one per intent. 

Age presents a telling reversal depending on intent. Users aged 55 and older 
are less likely to rely on GenAI chatbot for instrumental and enhancing tasks such as \textit{Content 
Creation} ($p < .001$) and \textit{Creativity} ($p < .01$). However, 
older individuals (65+) positively predict higher uptake of GenAI chatbots for \textit{Information Retrieval}.
%
%
Turning to competence variables, a clear divide emerges along a passive, recreational 
versus an active, instrumental axis. LT training is a strong and consistent 
positive predictor of \textit{Learning} ($p < .001$), \textit{Content Creation} ($p < .001$), and \textit{Creativity} ($p < .001$), while 
remaining non-significant for \textit{Information retrieval} and \textit{Leisure}. LT experience 
follows a similar pattern, positively predicting Content Creation and 
Creativity but not passive, recreational uses. LT literacy shows greater variation, reaching positive significance 
only for \textit{Problem Solving} and \textit{Content Creation}. 
\textit{Leisure} is uniformly insensitive to all competence measures. 
Taken together, these results suggest that higher uptake for more purposeful, enhancing uses are linked to training, while more passive, recreational usage remains accessible regardless of competence levels. 



Crucially, gender negatively predicts five of six intents. The largest gap appears in problem solving ($\beta = -0.38$, $p < .001$) and the smallest---and only non-significant--- in content creation ($\beta = -0.09$, $p < .10$). As such, being a woman emerges as the only systematically negative predictor in both first- and second-order divides, operating beyond competence, experience, and other sociodemographic controls.

\section{Discussion and Conclusion}\label{sec:discussion}

Our study provides a comprehensive, multi-layered empirical account of GenAI chatbot adoption, usage, and literacy among Italian-speaking adults, placing our work within a multilevel digital divide framework. Taken together, our findings attest to the key role of GenAI chatbots in the technological ecosystem. They are employed for many purposes and activities, spanning both personal and professional domains, and attracting a user base that already surpasses that of longstanding technologies such as voice assistants---sometimes replacing established tools like web search altogether. Yet, as of today, this diffusion is uneven.  
In the following, we conclude by discussing our key findings on first- and second-order divides and their implications. Building on these, we sketch implications for third-order differentials in outcomes and benefits. 

\subsection{The Multi-Level Divide: From Adoption to Literacy and Use}

A central contribution of this study is the joint examination of first- and second-order divides across a broad, general-population sample to show if and how social correlates of inequality operate at each level.
At the adoption level, familiar sociodemographic fault lines---gender, age, education, income---shape who crosses the threshold into use, consistent with other waves of digital technology diffusion \citep{francis2019aging}.\footnote{Notably, we find no regional differences. This null finding should be interpreted cautiously: our online survey sample equalizes internet access across regions, masking Italy's documented north-south digital divide \citep{SERRECCHIA2024102870}. Additionally, our analysis did not capture urban-rural distinctions, which concurrent research identifies as a meaningful predictor of GenAI adoption in the US \citep{zhou2025attentionnonadopters}, thus suggesting an important avenue for future investigation.}
Those with 
accumulated experience with other language technologies are more likely to adopt GenAI chatbots, in line with  research showing that greater technological familiarity drive greater acceptance \citep{Araujo2020}.
Once the adoption threshold is crossed, however, experiential and sociodemographic factors 
lose consistent influence and vary substantially across modes of use. 
We observe distinct patterns:  
purposeful, productive engagement is leveraged among those with access to training and greater skills.  In fact, prior training is associated with higher engagement for work or study contexts and more instrumental purposes. Thus, targeted education and competence are not merely a factor for whether one uses GenAI, but a lever for what one gets out of it---and one that is itself socially stratified, potentially compounding advantage across levels.

These findings have both empirical and methodological implications. Prior work relying solely on overall usage frequency or conflating adoption with actual engagement might miss critical variation. More substantively, our findings offer a first, more granular mapping of how engagement with GenAI chatbots is socially structured---a necessary step toward understanding who stands to benefit from them, and how.

Ultimately, these findings take on added weight in a country like Italy, where overall digital literacy levels are among the lowest in Europe \citep{EuropeanCommission2024Italy, ipsos2025aimonitor}. We find that one in two users has received no AI training, and self-assessed literacy remains low even among active adopters. The consequences are unlikely to be uniform: older users---already the least digitally literate---show the highest reliance on GenAI for retrieving factual information, despite being the least equipped to evaluate its outputs critically. 
%
%
Whether this might translate into third-order divides and concrete harms---susceptibility to errors, misinformation, or distorted political reasoning---lies beyond the scope of this study. Yet the structural conditions for such harms are already visible in our data, and documenting them represents a key avenue for future research.


\subsection{Non-adopters}
\label{subsec:non-adopters}

Another key contribution of this study is the inclusion of non-adopters---a group critically underrepresented in current research on GenAI chatbots. Human-Computer Interaction research  argues for taking non-use seriously on its own terms \citep{satchell}.
We give empirical substance to that call and identify two substantively distinct non-user groups that warrant different responses.
The first faces knowledge and competence barriers, which structured support could plausibly address. This points to a policy direction towards targeted GenAI literacy programs \citep{GenAIliteracy}, especially for older adults \citep{Ko_2025}.

The second is composed of deliberate, or \textit{voluntary} non-users \citep{wyatt2002they}, who perceive GenAI chatbots as lacking utility. 
Rather than treating this as a permanent state, we suggest it may reflect a considered response to a technology that does not serve their needs. 
Evidence from the United States lends support to this interpretation \citep{zhou2025attentionnonadopters}: GenAI chatbots' capabilities reflect the priorities of current users---writing, creativity, coding---while underserving tasks non-adopters value more, such as care planning, domestic tasks, and physical-world problem solving. 
This pattern aligns with ``me-search'' \citep{nash2011me}: a tendency for researchers and developers to inadvertently prioritize problems and aspects that reflect personal interests. 
The consequence is a self-reinforcing dynamic: models trained predominantly on the interactions of current users are iteratively optimized for their tasks and needs, rendering them progressively less relevant to non-adopters. 
From a third-order perspective, this dynamic suggests that the distribution of GenAI's benefits may increasingly reflect the composition of its current user base---disproportionately young, educated, and male---unless deliberate corrective action is taken at the level of design and evaluation \citep{bassignana-etal-2025-ai}.
%
Closing adoption gaps may therefore require not only literacy interventions but fundamental and broader attention to whose needs GenAI chatbots are designed to serve.

\subsection{Gender Gaps}

Gender emerges as the single factor that operates consistently and disadvantageously for women across
both first- and second-order divides in our data, surviving all controls. 
Besides a potential misalignment between women's priorities and current tasks supported by GenAI chatbots (see Section \ref{subsec:non-adopters}), trust, social norms, and the cultural meanings attached to technology are
likely at work \citep{wajcman2004techno, otis2024global}. 

The Italian context adds a specific structural dimension to this picture. 
Research on digital confidence suggests that the competence and familiarity that facilitate technology adoption tend to develop through informal, self-directed exploration. This process is more feasible in domestic settings than in pressurized work environments \citep{bracciale2010donne}. Italian women, however, bear a disproportionate share of caregiving and domestic work \citep{dondena2020}.\footnote{In 2020, 74\% of Italian women report they do not share household duties equally with male partners.}
This trend can help explain the generational pattern we observe: gender gaps in adoption are substantially larger among older cohorts, reflecting
the effect of time constraints over the life course. Whether younger women, who are more engaged than their predecessors, will sustain this advantage is a question our cross-sectional design cannot answer---but one we hope future work will address using the data we make publicly available.

One finding in our data resists easy interpretation and warrants focused future inquiry: despite using GenAI chatbots in fewer activities and less frequently overall, women mostly rely on them in work or study contexts. 
A plausible interpretation is that greater technology-related anxiety \citep{dondena2020} fosters a more cautious, instrumentally oriented engagement, crowding out recreational use. This is consistent with evidence from educational settings, where women are more likely to refrain from GenAI use out of concern about being perceived as cheating \citep{carvajal2024will}, suggesting that lower overall frequency may reflect perceived social risk. 

These patterns are not merely descriptive---they may already be translating into outcome differentials. For example, in academia, male researchers' publication productivity rose 6.4\% more than women's following the arrival of ChatGPT, a gap linked to higher male usage rates \citep{tang2025gender}. Yet the same evidence base points to an opportunity. When GenAI chatbots are permitted in educational settings, female students increase their use substantially, and acquiring GenAI chatbot competence has been shown to more strongly enhance women's labor market prospects \citep{carvajal2024will}. This suggests that the gender divide in GenAI engagement is not fixed---and that targeted policy interventions could not only close it, but leverage it as a vehicle for broader gender equity in the digital economy.

\printendnotes

\backmatter






\section*{Declarations}

\bmhead{Data availability} 
Data with participants' responses are available at \url{https://github.com/RiTA-nlp/Survey-GenAI-Italy}. The repository also contains the survey itself (original Italian and translated into English). 

\bmhead{Code availability}
All code used to prepare the data, analyze them, compute results and visualize them are 
in the same project repository \url{https://github.com/RiTA-nlp/Survey-GenAI-Italy}

\bmhead{Funding} 
The current work did not receive any supporting funding. 

\bmhead{Acknowledgements}
This work is the result of a joint effort of members of the ``Risorse per la Lingua Italiana'' community (\url{https://rita-nlp.org}): we thank every member who invested their time in the project.
Beatrice Savoldi, Matteo Negri and Luisa Bentivogli are supported by the Horizon Europe research and innovation programme, under grant agreement No 101135798, project Meetween (My Personal AI Mediator for Virtual MEETtings BetWEEN People). 
Giuseppe Attanasio was supported by the Portuguese Recovery and Resilience Plan through project C645008882-00000055 (Center for Responsible AI) and by FCT/MECI through contract UIDB/50008 (Instituto de Telecomunicações). Elisa Bassignana is supported by a research grant (VIL59826) from VILLUM FONDEN.
Arianna Muti and Debora Nozza are supported by the European Research Council (ERC) under the European Union’s Horizon 2020 research and innovation program (grant agreement No. 101116095, PERSONAE).
Nicoletta Balbo is supported by the European Union (ERC FRAILIFE—Child Disability and Family Life, G.A. 101077533) under the European Union’s Horizon 2021-2027 research and innovation program. 

\bmhead{Author Contribution}
 B.S. and G.A. conceived the original idea and designed the methodology, with revisions from E.B., S.C., A.R., M.N., L.B., T.C., N.B, and D.N. The responsibility for validation, formal analysis, and investigation was of B.S. with O.G. and G.A. Data curation was performed by B.S., G.A., M.M.M., S.C., and O.G. The original draft was written by B.S. with M.M.M. and G.A. Visualization was handled by B.S. and M.M.M. All authors reviewed and edited the final manuscript.

\bmhead{Conflict of interest} 
The author(s) declare no competing interests.

\section*{Ethics Statement}

\bmhead{Ethics approval}

This research project received ethical approval from the   Bocconi Research Ethics Committee (approval number EA000960)  through the expedited review procedure. The project was approved on 12 May 2025---before distributing the survey---and was deemed compliant with the Research Ethics Policy of Bocconi University. 
The approval covered the content of the survey (i.e., the questions administered to participants) and the data collection procedures, including the methods used to distribute the survey and recruit participants. All research procedures involving human participants were conducted in accordance with institutional and national ethical standards and the principles of the Declaration of Helsinki.

\bmhead{Consent to participate}

Informed consent to participate was obtained electronically for each participant before starting the survey. On the first page of the Qualtrics survey, participants were linked to the detailed informed consent document describing the purpose of the study, its voluntary nature, the procedures involved, data protection measures, and participants’ rights under the EU General Data Protection Regulation (GDPR 679/2016).
To proceed, participants were required to confirm the following written statement: \textit{I confirm that I am at least 18 years old and that I have received the provided information. I understand that my participation in the study is voluntary, anonymous, and unpaid, and that I may withdraw at any time. I acknowledge that my data will be processed in full compliance with applicable data protection regulations and I therefore consent to participate in this study, authorizing the use of my responses in aggregated form for research purposes.}
Only participants who actively selected \textit{Yes (I agree to participate in the study)} could access and complete the questionnaire. No personal identifiers were collected, and all responses were stored in anonymized form on secure servers accessible only to the research team.
The consent covered participation in the study, the use of anonymized data for research and publication in aggregated form, and data storage in compliance with relevant privacy regulations. No vulnerable individuals or minors were involved, and no deception or foreseeable risks were associated with participation.


\section*{Supplementary information}

We provide all supplementary content (Figures, Tables, and experimental details and analyses) as an additional PDF file.   


\listoffigures

\bigskip

\clearpage

\begin{center}
{\LARGE\bfseries Supplementary Materials}
\end{center}
\vspace{1em}


\renewcommand{\thefigure}{S\arabic{figure}}
\renewcommand{\thetable}{S\arabic{table}}
\renewcommand{\theequation}{S\arabic{equation}}
\renewcommand{\thesection}{S\arabic{section}}
\setcounter{figure}{0}
\setcounter{table}{0}
\setcounter{equation}{0}
\setcounter{section}{0}

\setcounter{section}{0}


\tableofcontents
\listoftables
\newpage

\section{Experimental Details}
\label{appA}

\subsection{LT and GenAI Chatbot Definitions}
\label{appA:definitions}

The definitions and examples of each language technology in our survey are provided in Table \ref{tab:definitions}. 

\begin{table}[htp]
\centering
\footnotesize
\small
\begin{tabular}{p{3.5cm}p{4.2cm}p{4.2cm}}
\toprule
\textbf{Item} & \textbf{Definition} & \textbf{Examples} \\
\midrule

\multicolumn{3}{l}{\textit{General Definition}} \\
\midrule

Language Technologies & Tools based on artificial intelligence (AI) used to analyze and produce human language, both written and spoken. & Voice assistants, machine translators, voice generation, assisted writing, speech transcript, chatbots  \\

\midrule
\multicolumn{3}{l}{\textit{Task-Specific LT Applications}} \\
\midrule

Machine Translation & Applications that automate the translation of text or speech from one language to another. & Google Translate, DeepL, Microsoft Bing, Amazon Translator \\

\addlinespace

Voice Assistants & Applications designed to perform tasks, provide information, and control devices through voice commands. & Siri, Google Assistant, Alexa, Cortana, Bixby, Celia \\

\addlinespace

Assisted Writing & Applications for text correction and revision (grammar, paraphrasing, and summarizing), or support for creative writing. It does not include generating text from scratch, only revision of already drafted text. & Grammarly, Writefull, QuillBot, Smart Compose (Google), Jasper AI \\

\addlinespace

Speech Transcription & Applications for transcribing audio content into text, such as dictation systems, subtitling tools, or applications for the hearing impaired. & Apple Note, Automatic transcription on Zoom / Google Meet / Teams, Otter.AI, Speechmatics, Rev \\

\addlinespace

Text-to-Speech & Applications for generating artificial voice and reading text aloud. & Google Cloud Text-to-speech, IBM Watson, ElevenLabs, Adobe Acrobat Read Aloud, Apple VoiceOver, NVDA, JAWS \\

\midrule
\multicolumn{3}{l}{\textit{General-Purpose GenAI  Systems}} \\
\midrule

Chatbots & General-purpose systems for creating text, images, audio, or other content from simple verbal prompts. Allow natural interaction via chat to obtain answers, support, or complete various tasks. & ChatGPT, Gemini, Claude, Copilot, Perplexity \\

\bottomrule
\end{tabular}
\caption{Definitions and examples of language technologies.}
\label{tab:definitions}
\end{table}

\subsection{List of GenAI Activities}
\label{appA:activities}

In the following, we provide a full list of the 24 GenAI activities (translated to English) included in the survey.

\begin{itemize}
    \item \textbf{Conversations and Support}
    \begin{itemize}
        \item Medical advice
        \item Personal advice
        \item Emotional or psychological support
        \item Other advice and suggestions
        \item Friendly conversations
        \item Philosophical or political conversations
        \item Romantic or erotic conversations
        \item Other conversations
    \end{itemize}
    
    \item \textbf{Generation and Content Creation}
    \begin{itemize}
        \item Quiz creation
        \item Audio or music generation
        \item Image generation
    \end{itemize}
    
    \item \textbf{Writing and Analysis}
    \begin{itemize}
        \item Analysis (general)
        \item Data analysis and comprehension
        \item Programming language assistance
        \item Summaries
        \item Academic writing
        \item Creative writing
        \item Email writing
        \item Other writing
    \end{itemize}
  \item \textbf{Others}  
  \begin{itemize}
          \item Travel and itinerary planning
        \item Idea collection and draft creation
         \item Explanations
        \item Text comprehension and formatting
         \item Fact checking (verification of information accuracy)
          \end{itemize}
        
\end{itemize}

\subsection{Demographic Sample}\label{appAsample}

\begin{figure}[htp]
    \centering
    \includegraphics[width=1\linewidth]{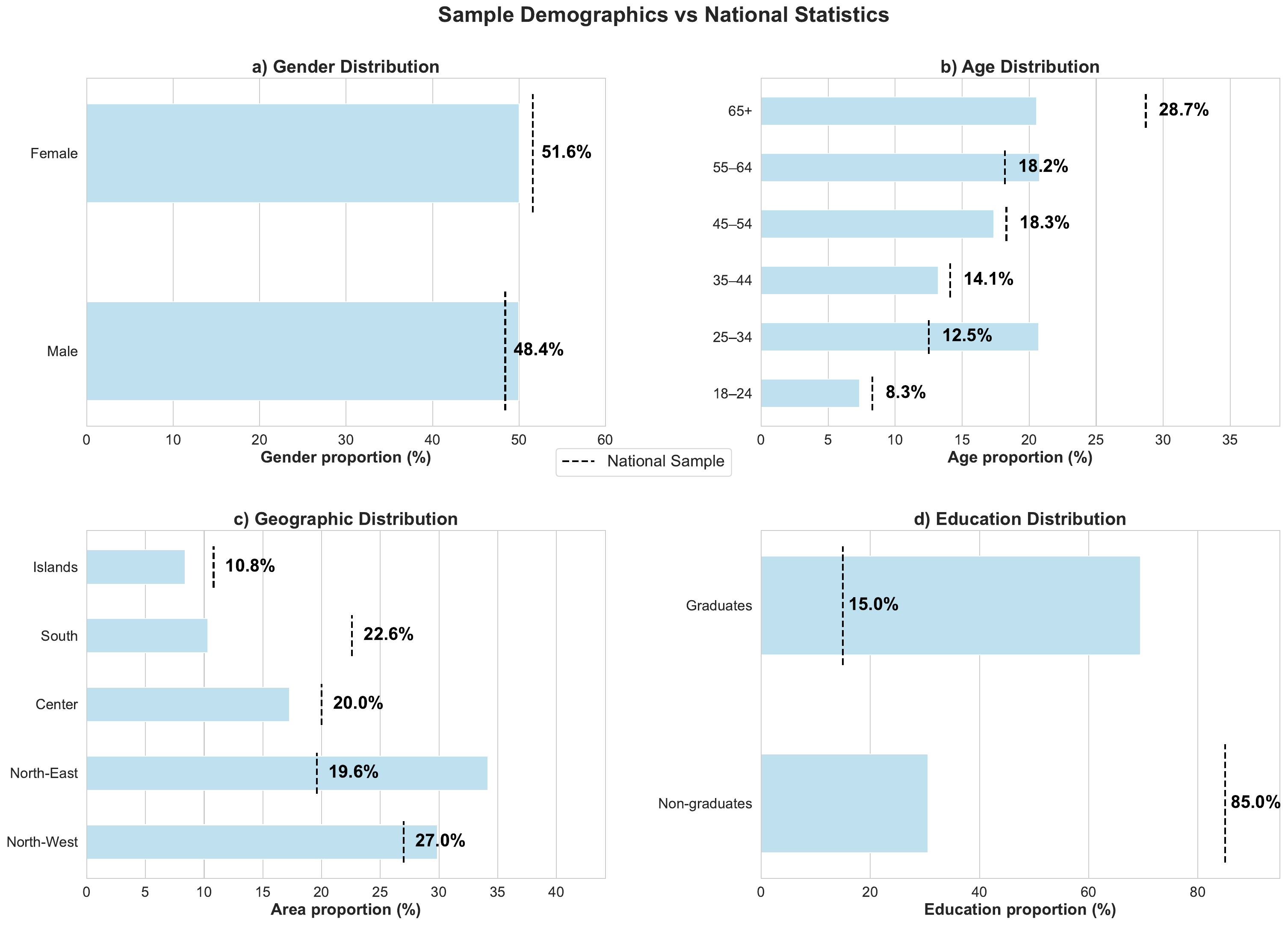}
    \caption{\textbf{Distribution of demographic characteristics in our sample} (teal bars) relative to national statistics (black dashed reference lines), broken down by (\textbf{a}) gender, (\textbf{b}) age, (\textbf{c}) geographic area, and (\textbf{d}) education.}
    \label{fig_app:demographics_comparison}
\end{figure}

To compare our demographic sample, in Figure \ref{fig_app:demographics_comparison} we plot the population of our survey (teal bars) compared to the national one (red dashed reference lines). 
For representative values and distribution, we refer to the Italian National Institute of Statistics
(ISTAT) 2025 data (\url{https://demo.istat.it/app/?i=POS&l=it}).
Values are reported as percentages across four key demographic categories, i.e., gender, age, geographic area, and education.
Robustness checks on our sample are reported in \ref{appB:robustness}.



\begin{figure}[htp]
  \centering
  \begin{promptblock}
I wrote and published a survey to understand how Italians use generative AI. The survey and the participant responses are in Italian.
Now, I am analyzing the following open-form question:

# Question
[question]

After clustering my answers, I have identified a topic described by the following keywords. 

# Keywords
[keywords]

The following responses are a representative sample of the cluster representing the topic

# Documents
[documents]

# Guidelines
Based on this information, your task it to generate a topic description in the context of this question. Here are the guidelines:
1. The description must be detailed and exhaustive, but it CANNOT exceed 300 words.
2. Describe what brought you to the description by mentioning the sample answers. 
3. Do not mention the question, it's known to the reader. Moreover, do not make any assumption of the number of responses in the cluster.
4. Begin the description with a short title that summarizes it, e.g., "Enhanced perceived precision. <description>", or "Chosen due to convenience. <description>"
5. If the answers are simply "no" or "nessuna", this is a cluster with negative answers. In this case, just avoid spending many words and say "Negative answers". 

Begin the answer right after this prompt.
  \end{promptblock}
  \caption{\textbf{Cluster summarization prompt.} [question], [keywords], and [documents] are replaced with one of the questions of the survey, the 10 most relevant keywords and documents describing the cluster, respectively.}
    \label{lst:cluster-prompt}
\end{figure}

\subsection{BERTopic Details and Configuration} \label{ssec:app:bertopic}

BERTopic is Python software\footnote{\url{https://maartengr.github.io/BERTopic/index.html}} commonly used for topic modeling. Off-the-shelf, it offers an integrated interface to a wide range of natural language processing tools, along with high customizability.

Our BERTopic processing pipeline includes several steps.
First, it converts each text individually into a numerical representation. For this step, we used a state-of-the-art \emph{sentence embedder} model, namely Qwen3-Embedding-8B \citep{qwen3embedding}. Sentence embedders capture semantic meaning in a format suitable for computational analyses by computing an N-dimensional vector representation for the text, also known as \emph{embedding}. The geometrical properties of embeddings are commonly used for semantic search and information retrieval (e.g., the cosine similarity between embeddings is often used to retrieve documents relevant to a user query).
Second, BERTopic applies dimensionality reduction using UMAP \citep{mcinnes2018umap} to minimize noise by reducing the embedding dimensionality.
Third, it uses the HDBSCAN clustering algorithm \citep{mcinnes2017hdbscan} to group semantically similar responses. In all cases, this step produced no more than 25 distinct clusters.
Finally, for each cluster, the algorithm selects the N most relevant texts and compiles a prompt for a large language model. For this step, we prompted Gemma 3 Instruct \citep{gemmateam2025gemma3technicalreport}, a leading generative language model, to provide a detailed but concise summary of the cluster’s topics. Listing~\ref{lst:cluster-prompt} reports the prompt we used.
Since some textual nuances introduce noise, we manually verified all extracted clusters and merged similar ones. 



Before running the BERTopic pipeline, we exclude all responses shorter than three characters and remove a set of Italian stopwords as found at \url{https://github.com/stopwords-iso/stopwords-it/blob/master/stopwords-it.txt} from the remaining ones. We manually tuned two UMAP and HDBSCAN parameter sets to adjust cluster granularity and diversity based on the question. Table~\ref{tab:hparams-bertopic} reports all the final parameters.

\begin{table}[htp]
\centering
\begin{tabular}{@{}llrr@{}}
\toprule
\textbf{Method} & \textbf{Parameter} & \multicolumn{1}{l}{\textbf{Language Choice}} & \multicolumn{1}{l}{\textbf{Modality Type}} \\ \midrule
UMAP & n\_neighbors & 10 & 5 \\
 & n\_components & 20 & 20 \\ \midrule
HDBSCAN & min\_cluster\_size & 10 & 6 \\
 & min\_samples & 5 & 5 \\ \bottomrule
\end{tabular}
\caption{\textbf{BERTopic parameters.} Final choice for UMAP and HDBSCAN parameters for the open-text questions about users' expressed preference about the language chosen to prompt the model.}
\label{tab:hparams-bertopic}
\end{table}



\section{Additional Results}\label{app:B}

\subsection{Sample Robustness and Sensitivity}
\label{appB:robustness}

To assess the robustness of our findings, we applied post-stratification weights based on Italian population benchmarks from ISTAT 2025 data for gender, age, and geography. We computed frequency tables comparing unweighted and weighted percentages across all key variables, calculating the delta ($\Delta$; difference in percentage points) for each categorical response. We classified variables using three thresholds based on the maximum delta across categories: \textit{pass} ($\Delta \leq 3$ pp), \textit{review} ($3 < \Delta \leq 5$ pp), and \textit{critical} ($\Delta > 5$ pp). We also verified whether the top-2 responses remained consistent between weighted and unweighted distributions.

\begin{figure}[htp!]
    \centering
    \includegraphics[width=1\linewidth]{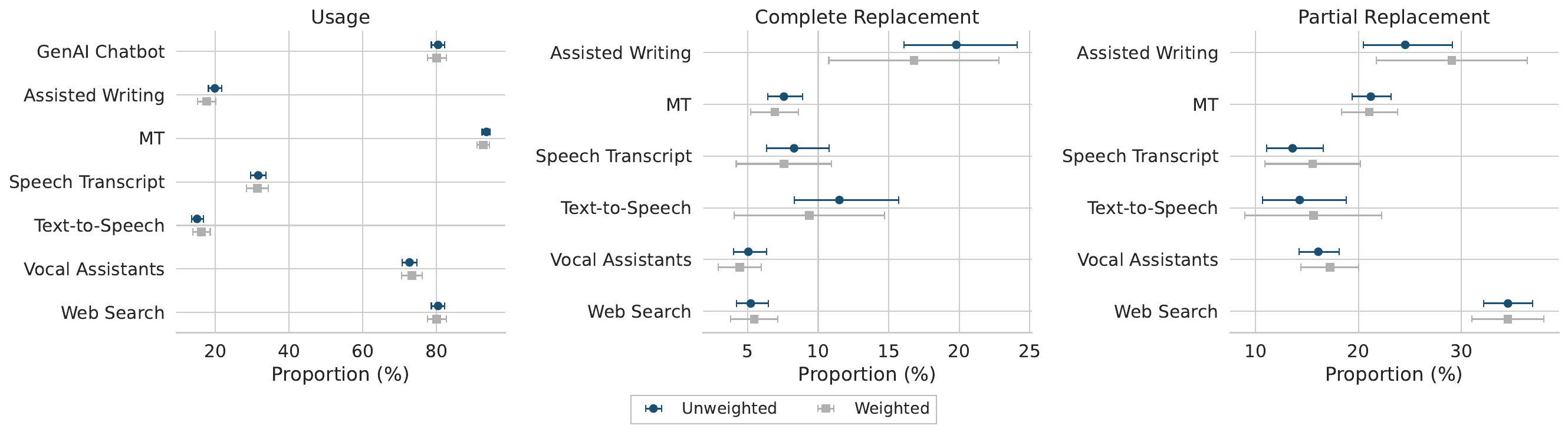}
    \caption{\textbf{Sample robustness.} Forest plot comparing unweighted and weighted proportion estimates across three panels: technology usage (\textit{left}), complete replacement (\textit{center}), and partial replacement (\textit{right}) of pre-existing language technologies. Blue filled circles represent unweighted estimates; gray filled circles represent post-stratification weighted estimates (ISTAT 2025). Error bars indicate 95\% confidence intervals.}
    \label{figure:robustness_ci}
\end{figure}

    Key variables showed minimal weighting effects. Educational level and socioeconomic status all passed the robustness check with deltas under 2 pp. GenAI chatbot adoption (80.4\% vs.\ 80.1\% weighted), language technology usage and replacement behaviors were similarly robust. In Figure \ref{figure:robustness_ci} we present a forest plot comparing unweighted and weighted proportion estimates with 95\% confidence intervals across technology usage and replacement of pre-existing language technologies. 
    %
    For all technology measure combinations, weighted and unweighted 95\% CIs overlap.

AI Training distribution, usage modalities, strategies, and prior experiences with errors also showed stable patterns. Intent frequencies for using GenAI passed checks with deltas under 2.5 pp, with the sole exception of the ``Learning'' intent, where weighted estimates showed slightly higher weekly use ($\Delta = 3.02$ pp).

All substantive directions remained consistent, and only two sets of responses exhibited larger weighting effects. 
English language use showed a $5.06$ pp delta, with weighted samples showing lower usage (48.4\% vs 51.6\%), potentially reflecting over-representation of younger respondents. 
Work/personal usage by income subgroup showed borderline weighting effects: Income: Lower ($n = 270$, $\Delta = 3.73$ pp, \textit{review}) and Income: Higher ($n = 437$, $\Delta = 3.10$ pp, \textit{review}). Both fall within the \textit{review} threshold ($3 < \Delta \leq 5$ pp) and are not included in the main text analysis.


Overall, our main findings regarding demographic disparities, adoption and engagement patterns remain robust to post-stratification weighting, supporting our decision to report unweighted results and retain all participant responses.

\subsection{LT Training and Literacy Moderation Analysis}
\label{app:ssec:training_literacy_moderation_analysis}

We test two hypotheses: first, that LT training is positively associated 
with self-assessed LT literacy; second, that this association is moderated by gender. 

Table~\ref{tab:training_literacy} reports  
reports OLS regression models in which 
self-assessed LT literacy is the dependent variable, LT training is the 
independent variable, and sociodemographic characteristics (age, gender, 
income, education, and geography) are included as controls, estimated 
separately for chatbot users and non-users.
LT training is significantly associated with 
self-assessed LT literacy in both subsamples (users: $\beta = 0.133$, 
$p < .001$; non-users: $\beta = 0.099$, $p < .001$), confirming the first 
hypothesis.


\begin{table*}[htp!]
\centering
\small
\begin{tabular}{lll}
\toprule
 & \textbf{Non-users} & \textbf{Users} \\
\midrule
Intercept & 0.425*** (0.045) & 0.569*** (0.017) \\
\textbf{LT Training} & \textbf{0.099***} (0.027) & \textbf{0.133***} (0.010) \\
Woman & -0.014 (0.024) & -0.055*** (0.010) \\
Age 35–54 & -0.016 (0.042) & -0.022 (0.013) \\
Age 55–64 & -0.092* (0.041) & -0.041** (0.014) \\
Age 65+ & -0.057 (0.040) & -0.107*** (0.016) \\
Graduates & 0.033 (0.025) & 0.005 (0.012) \\
Income Mid & -0.045 (0.029) & -0.003 (0.014) \\
Income Higher & 0.006 (0.037) & 0.013 (0.015) \\
Geography  Centre & 0.014 (0.032) & -0.016 (0.014) \\
Geography  South \& Islands & 0.030 (0.032) & 0.001 (0.014) \\
Geography  Abroad & 0.175** (0.064) & 0.046* (0.021) \\
\midrule
$R^2$ & 0.125 & 0.168 \\
$N$ & 363 & 1,510 \\
\hline \\[-1.8ex]
\textit{Note:} & \multicolumn{2}{r}{$^{*}$p$<$0.05; $^{**}$p$<$0.01; $^{***}$p$<$0.001} \\
\bottomrule
\end{tabular}
\caption{\textbf{OLS regression models using LT training and sociodemographic controls to predict LT literacy}, estimated separately for chatbot users and non-users. All coefficients shown. Standard errors in parentheses.}
\label{tab:training_literacy}
\end{table*}

Table~\ref{tab:training_literacy_gender} reports a moderation analysis 
estimated via OLS regression, with self-assessed LT literacy as the dependent 
variable, LT training as the independent variable, gender as the moderator, and sociodemographic controls (age, 
income, education, and geography). 


\begin{table*}[htp!]
\centering
\small
\begin{tabular}{lll}
\toprule
 & \textbf{Non-users} & \textbf{Users} \\
\midrule
Intercept & 0.433*** (0.046) & 0.551*** (0.018) \\
LT Training & 0.067 (0.044) & 0.162*** (0.014) \\
Woman & -0.027 (0.027) & -0.025 (0.014) \\
Age 35–54 & -0.014 (0.042) & -0.022 (0.012) \\
Age 55–64 & -0.093* (0.041) & -0.038** (0.014) \\
Age 65+ & -0.058 (0.040) & -0.103*** (0.016) \\
Graduates & 0.033 (0.025) & 0.005 (0.012) \\
Income Mid & -0.042 (0.030) & -0.001 (0.014) \\
Income Higher & 0.006 (0.037) & 0.014 (0.015) \\
Geography  Centre & 0.012 (0.032) & -0.016 (0.014) \\
Geography   South \& Islands & 0.027 (0.032) & 0.003 (0.014) \\
Geography  Abroad & 0.175** (0.064) & 0.047* (0.021) \\
\textbf{LT Training × Woman} & 0.050 (0.055) &\textbf{ -0.062**} (0.020) \\
\midrule
$R^2$ & 0.127 & 0.173 \\
$N$ & 363 & 1,510 \\
\hline \\[-1.8ex]
\textit{Note:} & \multicolumn{2}{r}{$^{*}$p$<$0.05; $^{**}$p$<$0.01; $^{***}$p$<$0.001} \\
\bottomrule
\end{tabular}
\caption{Gender moderation of the LT training effect on LT literacy, estimated separately for chatbot users and non-users. All coefficients shown. Standard errors in parentheses.}
\label{tab:training_literacy_gender}
\end{table*}

Among chatbot users, the gender variable is not statistically meaningful 
 (Woman $\beta = -0.025$, n.s.), suggesting no baseline 
gender gap among untrained men and women. Training positively predicts LT literacy ($\beta = 0.162$, $p < .001$), but the interaction term (LT Training $\times$ Woman $\beta = -0.062$, $p < .01$) indicates that this benefit is significantly attenuated for women, thus confirming the second hypothesis. Among 
non-users, neither the gender variable nor the interaction term 
reach statistical significance.  



\subsection{Multi-collinearity Verification}
\label{app:ssec:multicollinearity}


\begin{table}[htp]
\centering
\begin{tabular}{lrrr}
\toprule
\textbf{Variable} & \textbf{VIF} & \textbf{Df} & \textbf{VIF (Norm)} \\
\midrule
Age Group & 1.179 & 1 & 1.086 \\
Education Group & 1.086 & 1 & 1.042 \\
Gender Group & 1.033 & 1 & 1.016 \\
Geography Group & 1.017 & 2 & 1.004 \\
Income Group& 1.074 & 1 & 1.036 \\
LT Experience & 1.139 & 1 & 1.067 \\
LT Training & 1.152 & 1 & 1.073 \\
LT Literacy & 1.287 & 1 & 1.134 \\
\bottomrule
\end{tabular}
\caption{\textbf{Variance Inflation Factors (VIF).} Values for each predictor variable in our study, degree of freedom (Df), and normalized VIF (Norm) obtained as $\text{VIF}^{1/2 \cdot \text{Df}}$.}
\label{tab:VIF}
\end{table}

Multi-collinearity between predictor variables may hinder the reliability of regression analyses. Across our experiments, we use eight variables divided into numerical, ordinal, and categorical: Education Group (categorical), Geography Group (categorical), Gender Group (categorical), Socioeconomic Group (ordinal), Age Group (ordinal), and LT Training (categorical), LT Experience (numerical), and LT Literacy (numerical).

To verify multi-collinearity, we follow a standard approach based on Variance Inflation Factors \citep[VIF;][]{o2007caution}. Specifically, we use linear regression and $R^2$ to estimate numerical variables and logistic-type models and McFadden's pseudo-$R^2$ for categorical and ordinal ones.
Table \ref{tab:VIF} reports the VIF value for each variable. The highest score is 1.287, indicating a non-problematic multicollinearity.

\subsection{Complementary Descriptive Results and Visualizations}
\label{app:ssec:complementary_results}

Complementary descriptive results and visualizations.

\begin{figure}[htp!]
    \centering
    \includegraphics[width=1\linewidth]{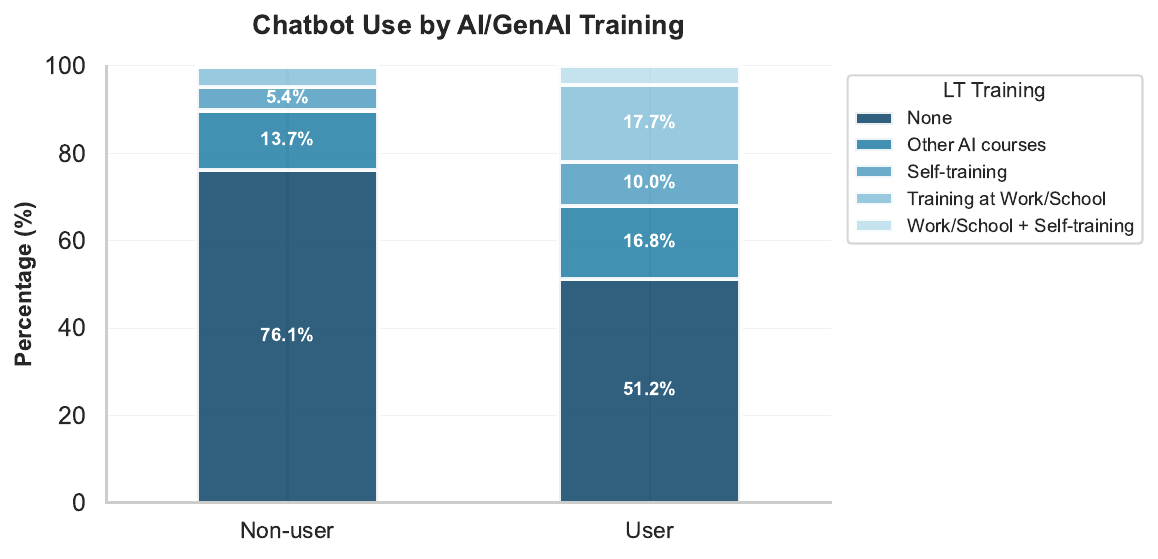}
    \caption{\textbf{Prior LT training by GenAI chatbot usage status.}}
\label{fig:training_by_usage}.
    \label{fig_app:prior_training}
\end{figure}

Figure \ref{fig_app:prior_training} shows prior \textbf{LT training background} by chatbot usage status, highlighting that non-users are less likely to have received any formal or informal training related to GenAI chatbot or AI and AI language technologies more broadly.  Respondents were asked whether they had received prior training in AI-based LT technology via both formal or informal channels. Among chatbot users, 51.2\% reported no prior training, while 76.2\% of non-users had no training background at all. Users showed higher rates of formal training through work/school (16.8\% vs. 0\%), as well as combined formal and self-training (17.6\% vs. 5.2\%). A small proportion in both groups reported other AI-related coursework (User: 10.1\%, Non-user: 13.6\%).

\begin{table*}[htp!]
\small
\begin{tabular}{lrrrrrrr}
\toprule
 & U-M & U-SD & NU-M & NU-SD & t-stat & p-value & Cohen-d \\
\midrule
\textbf{LT Literacy} & & &&&&&\\
\quad Distinguish AI & -0.027 & 1.082 & -0.276 & 1.247 & 3.861 & 0.000 & 0.223  \\
\quad Recognize errors & 0.417 & 1.075 & -0.303 & 1.290 & 11.143 & 0.000 & 0.643  \\
\quad Potential & 0.373 & 1.152 & -0.501 & 1.226 & 12.977 & 0.000 & 0.749  \\
\quad Limitations & 0.299 & 1.209 & -0.550 & 1.338 & 11.895 & 0.000 & 0.687  \\
 \quad  Prepared & 0.234 & 1.178 & -0.842 & 1.124 & 15.959 & 0.000 & 0.921  \\
 \quad Knowledge & 0.061 & 1.142 & -0.992 & 1.081 & 16.145 & 0.000 & 0.932  \\
\midrule
LT Education & 1.172 & 1.083 & 0.507 & 1.377 & 10.048 & 0.000 & 0.580  \\
Bias awareness & 0.965 & 1.154 & 0.560 & 1.320 & 5.896 & 0.000 & 0.340  \\
\bottomrule
\end{tabular}
\caption{\textbf{Descriptive statistics for LT literacy items, LT education attitude, and bias awareness by GenAI chatbot usage status.} U = users; NU = non-users. Means (M) and standard deviations (SD) are reported for each group on the recoded scale (ranging from $-$2 to $+$2). Cohen's $d$ is reported for effect size. Statistical significance measured with independent samples $t$-tests.}
\label{table:lit-stats}
\end{table*}

Table \ref{table:lit-stats} reports the \textbf{full descriptive statistics (means and standard deviations) for each LT literacy item, LT education attitude, and bias awareness by GenAI chatbot usage status}, alongside independent samples $t$-test statistics and Cohen's $d$ effect sizes. Effect sizes range from small ($d = 0.22$ for distinguishing AI from human text) to large ($d = 0.93$ for self-reported LT knowledge), confirming that GenAI users and non-users differ substantially across all LT literacy dimensions.

\begin{figure}[htp!]
    \centering
    \includegraphics[width=1\linewidth]{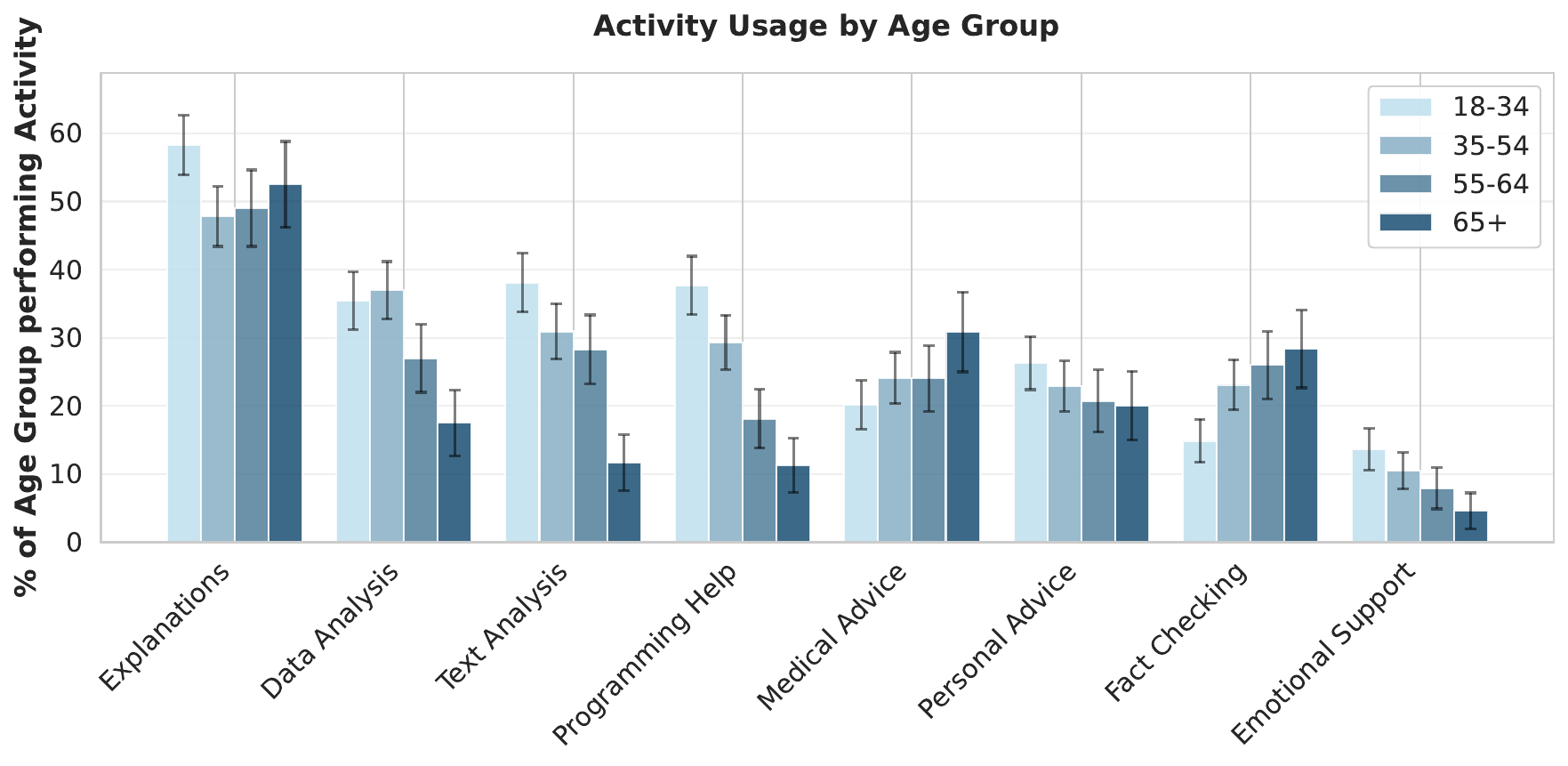}
    \caption{\textbf{GenAI Chatbot activity by age groups}. Bars show the percentage of users in each age group performing the activities, with 95\% confidence intervals.}
    \label{fig:activities_age}
\end{figure}

Figure \ref{fig:activities_age} reports the percentage of users across age groups that perform each of the shown \textbf{activities with GenAI chatbots}. While variation is moderate for seeking explanations or personal advice, younger users predictably dominate technical tasks such as data analysis, text analysis, and programming help. Notably, older users (65+) stand out for seeking medical advice---likely driven by healthcare needs---and fact-checking, despite these models' established susceptibility to hallucinations and misinformation \citep[e.g.][]{kreps2022all, zhang2025siren, li-etal-2024-dawn}.


    

\subsection{Usage Frequency by Intent Modeling}
\label{app_subsec:usage_frequency}

Table \ref{tab:ols_intent} shows \textbf{OLS regression coefficients} predicting the frequency of use of GenAI chatbots by intent, estimated separately for each of the six use intents (columns). 
While usage frequency is formally an ordered measure (0--3 scale), it can be reasonably interpreted as approximating a continuous variable, justifying the use of OLS for parsimony and ease of interpretation across models. To verify this choice, we estimated equivalent ordered logit models for each intent and compared the results. 

Table \ref{tab:ologit_intent} reports the \textbf{ordered logit coefficients} as a robustness check. The two specifications corroborate the same results: the Pearson correlation between OLS and ordered logit coefficients across all six models is $r = 0.993$ ($p < .001$), with a sign agreement of 96.2\% and a significance sign agreement of 98.7\%.  
The three sign disagreements involve coefficients close to zero (\textit{LT literacy} and \textit{LT training} in Information Retrieval, and \textit{Income Mid} in Content Creation), and the one significance disagreement involves a borderline $p$-value 
straddling the 0.05 threshold (\textit{Geography Centre} in Leisure). Neither affects any substantive conclusion.

The analytical sample is restricted to GenAI chatbot users only ($n = 1{,}533$; see Table 2 in the main paper), from which 23 individuals who identified their gender as ``Neither'' were excluded, resulting in a base sample of 1,510 respondents. Sample sizes vary slightly across the six models (range: 1,381--1,438) due to listwise deletion of missing values on the outcome variable for each intent.

\begin{table}[t!]
\centering
\small
\begin{tabular}{lrrrrrr}
\toprule
 & \textbf{Information } & \textbf{Problem} & \textbf{Learning} & \textbf{Content} & \textbf{Leisure} & \textbf{Creativity} \\
\midrule
\textbf{Gender} \\
\quad Woman & -0.219*** & -0.383*** & -0.288*** & -0.094 & -0.298*** & -0.148** \\
 & (0.060) & (0.057) & (0.056) & (0.057) & (0.056) & (0.057) \\
\textbf{Age} \\
\quad 35--54 & 0.144* & -0.024 & -0.083 & 0.078 & -0.027 & -0.044 \\
 & (0.073) & (0.070) & (0.068) & (0.069) & (0.067) & (0.069) \\
\quad 55--64 & 0.186* & -0.081 & -0.207** & -0.186* & -0.007 & -0.207* \\
 & (0.085) & (0.081) & (0.080) & (0.080) & (0.079) & (0.080) \\
\quad 65+ & 0.328*** & 0.024 & -0.052 & -0.628*** & -0.019 & -0.245** \\
 & (0.097) & (0.094) & (0.090) & (0.092) & (0.092) & (0.093) \\
\textbf{Education} \\
\quad Graduates & 0.034 & -0.051 & -0.110 & 0.217*** & -0.113 & -0.104 \\
 & (0.068) & (0.065) & (0.064) & (0.065) & (0.064) & (0.064) \\
\textbf{Geography} \\
\quad Abroad & -0.052 & 0.156 & 0.050 & 0.241* & -0.108 & 0.008 \\
 & (0.123) & (0.117) & (0.115) & (0.116) & (0.114) & (0.116) \\
\quad Centre & 0.098 & 0.116 & 0.034 & 0.050 & 0.133 & 0.133 \\
 & (0.082) & (0.078) & (0.076) & (0.077) & (0.076) & (0.077) \\
\quad South \& Islands & -0.034 & 0.001 & 0.062 & 0.109 & 0.140 & -0.041 \\
 & (0.080) & (0.077) & (0.075) & (0.075) & (0.075) & (0.076) \\
\textbf{Income} \\
\quad Higher & 0.236** & 0.096 & 0.266** & 0.203* & -0.072 & 0.197* \\
 & (0.090) & (0.087) & (0.085) & (0.085) & (0.083) & (0.086) \\
\quad Mid & 0.054 & 0.036 & 0.072 & 0.010 & -0.170* & 0.013 \\
 & (0.081) & (0.077) & (0.076) & (0.077) & (0.074) & (0.077) \\
\textbf{LT Experience} & 0.029 & 0.063* & 0.013 & 0.127*** & 0.040 & 0.128*** \\
 & (0.029) & (0.028) & (0.027) & (0.027) & (0.027) & (0.027) \\
\textbf{LT Literacy} & -0.009 & 0.336* & 0.176 & 0.704*** & 0.231 & 0.180 \\
 & (0.154) & (0.146) & (0.143) & (0.144) & (0.143) & (0.145) \\
\textbf{LT Training} & 0.004 & 0.096 & 0.250*** & 0.284*** & 0.071 & 0.291*** \\
 & (0.063) & (0.061) & (0.059) & (0.060) & (0.059) & (0.060) \\
Intercept & 1.836*** & 1.763*** & 1.996*** & 0.953*** & 0.763*** & 0.977*** \\
 & (0.145) & (0.138) & (0.135) & (0.136) & (0.134) & (0.136) \\
\midrule
Observations & 1438 & 1416 & 1433 & 1418 & 1381 & 1416 \\
$R^2$ & 0.034 & 0.056 & 0.053 & 0.168 & 0.044 & 0.071 \\
Adjusted $R^2$ & 0.025 & 0.048 & 0.045 & 0.161 & 0.034 & 0.063 \\
\bottomrule
\hline
\hline \\[-1.8ex]
\textit{Note:} & \multicolumn{6}{r}{$^{*}$p$<$0.05; $^{**}$p$<$0.01; $^{***}$p$<$0.001} \\

\end{tabular}
\caption{\textbf{OLS Regression Models: GenAI Chatbot Usage Frequency by Intent}. We run six separate OLS models, one per intent (columns). Coefficients and Standard Errors (in parentheses) are reported.}
\label{tab:ols_intent}
\end{table}

\begin{table}[t!] \centering
\begin{tabular}{@{\extracolsep{5pt}}lcccccc}
\\[-1.8ex]\hline
\hline \\[-1.8ex]
\\[-1.8ex] & \multicolumn{1}{c}{\textbf{Information}} & \multicolumn{1}{c}{\textbf{Problem }} & \multicolumn{1}{c}{\textbf{Learning}} & \multicolumn{1}{c}{\textbf{Content}} & \multicolumn{1}{c}{\textbf{Leisure}} & \multicolumn{1}{c}{\textbf{Creativity}}  \\
\hline \\[-1.8ex]
\textbf{Gender} \\
\quad Woman & -0.389*** & -0.696*** & -0.512*** & -0.137 & -0.641*** & -0.233* \\
 & (0.103) & (0.102) & (0.103) & (0.104) & (0.111) & (0.099) \\
\textbf{Age} \\
\quad 35--54 & 0.268* & -0.002 & -0.115 & 0.153 & -0.112 & -0.043 \\
 & (0.124) & (0.124) & (0.125) & (0.127) & (0.130) & (0.119) \\
\quad 55--64 & 0.343* & -0.123 & -0.337* & -0.321* & -0.128 & -0.374** \\
 & (0.145) & (0.143) & (0.145) & (0.145) & (0.156) & (0.141) \\
\quad 65+ & 0.527** & 0.053 & -0.092 & -1.046*** & -0.108 & -0.439** \\
 & (0.168) & (0.167) & (0.166) & (0.166) & (0.184) & (0.163) \\
\textbf{Education} \\
\quad Graduates & 0.042 & -0.085 & -0.194 & 0.349** & -0.173 & -0.180 \\
 & (0.117) & (0.116) & (0.118) & (0.117) & (0.125) & (0.113) \\
\textbf{Geography} \\
\quad Abroad & -0.089 & 0.307 & 0.025 & 0.520* & -0.199 & 0.009 \\
 & (0.208) & (0.212) & (0.205) & (0.233) & (0.228) & (0.203) \\
\quad Centre & 0.185 & 0.236 & 0.073 & 0.117 & 0.290* & 0.221 \\
 & (0.142) & (0.142) & (0.140) & (0.141) & (0.145) & (0.136) \\
\quad South \& Islands & -0.029 & 0.019 & 0.185 & 0.148 & 0.250 & -0.068 \\
 & (0.139) & (0.136) & (0.141) & (0.137) & (0.146) & (0.131) \\
\textbf{Income} \\
\quad Higher & 0.395* & 0.188 & 0.468** & 0.361* & -0.163 & 0.353* \\
 & (0.156) & (0.154) & (0.158) & (0.157) & (0.161) & (0.151) \\
\quad Mid & 0.068 & 0.072 & 0.101 & -0.009 & -0.298* & 0.043 \\
 & (0.137) & (0.136) & (0.138) & (0.139) & (0.143) & (0.134) \\
\textbf{LT Experience} & 0.050 & 0.121* & 0.021 & 0.241*** & 0.097 & 0.233*** \\
 & (0.049) & (0.050) & (0.050) & (0.051) & (0.052) & (0.048) \\
\textbf{LT Literacy} & 0.078 & 0.587* & 0.355 & 1.299*** & 0.359 & 0.311 \\
 & (0.267) & (0.266) & (0.265) & (0.263) & (0.281) & (0.257) \\
\textbf{LT Training} & -0.001 & 0.181 & 0.392*** & 0.519*** & 0.218 & 0.497*** \\
 & (0.109) & (0.107) & (0.109) & (0.109) & (0.117) & (0.105) \\
\hline \\[-1.8ex]
\multicolumn{7}{l}{\textit{Thresholds}} \\
 0.0/1.0 & -1.419*** & -1.521*** & -1.936*** & 0.025 & 0.299 & -0.376 \\
 & (0.254) & (0.251) & (0.259) & (0.250) & (0.260) & (0.244) \\
 1.0/2.0 & 0.025 & 0.197*** & 0.172** & 0.056 & 0.075 & 0.338*** \\
 & (0.061) & (0.057) & (0.061) & (0.061) & (0.054) & (0.043) \\
 2.0/3.0 & -0.286*** & -0.027 & -0.050 & 0.007 & -0.349*** & 0.176*** \\
 & (0.059) & (0.052) & (0.053) & (0.053) & (0.087) & (0.050) \\
\hline \\[-1.8ex]
 Observations & 1438 & 1416 & 1433 & 1418 & 1381 & 1416 \\
 AIC & 3495.494 & 3545.684 & 3471.288 & 3431.673 & 3016.469 & 3821.589 \\
 BIC & 3579.830 & 3629.774 & 3555.568 & 3515.785 & 3100.158 & 3905.679 \\
 McFadden $R^2$ & 0.014 & 0.025 & 0.020 & 0.068 & 0.022 & 0.027 \\
\hline
\hline \\[-1.8ex]
\textit{Note:} & \multicolumn{6}{r}{$^{*}$p$<$0.05; $^{**}$p$<$0.01; $^{***}$p$<$0.001} \\
\end{tabular}
\caption{\textbf{Ordered Logit Regression Models: GenAI Usage Frequency by Intent}. We run six separate ordered logit models, one per intent (columns). Coefficients and Standard Errors (in parentheses) are reported.}
  \label{tab:ologit_intent}
\end{table}

\end{document}